\documentclass{sig-alternate}
\usepackage{times}
\usepackage[numbers]{natbib}
\usepackage{verbatim}
\usepackage{amsmath}
\usepackage{amssymb}
\usepackage{url}
\usepackage{array}
\usepackage{booktabs}
\usepackage{afterpage}
\usepackage[dvipsnames]{color}
\usepackage{mfirstuc}
\makeatletter
\newif\if@restonecol
\makeatother

\usepackage[lined,algonl,boxed]{algorithm2e}%
\usepackage{epsfig}
\usepackage{multirow}

\begin{document}
\conferenceinfo{KDD'11,} {August 21--24, 2011, San Diego, California, USA.}
\CopyrightYear{2011}
\crdata{978-1-4503-0813-7/11/08}
\clubpenalty=10000
\widowpenalty = 10000

\newcommand{\vpara}[1]{\vspace{0.05in}\noindent\textbf{#1 }}
\newcommand{\para}[1]{\vspace{0.05in}\noindent\textbf{#1 }}
\newcommand{\secref}[1]{Section~\ref{#1}} 
\newcommand{\Real}{\ensuremath{\mathbb{R}}}  
\newcommand{\figref}[1]{Figure~\ref{#1}} 
\newcommand{\beq}[1]{\vspace{-0.02in}\small\begin{equation}#1\end{equation}\vspace{-0.02in}\normalsize}
\newcommand{\beqn}[1]{\vspace{-0.03in}\small\begin{eqnarray}#1\end{eqnarray}\vspace{-0.03in}\normalsize}
\newcommand{\besp}[1]{\begin{split}#1\end{split}}

\newcommand{\llee}[1]{\textcolor{red}{#1}}
\newcommand{\chenhao}[1]{\textcolor{blue}{#1}}

\newcommand{\follow}{\mbox{t-follow}\xspace}
\newcommand{\follows}{{\follow}s\xspace}
\newcommand{\Follow}{\mbox{t-Follow}\xspace}

\newcommand{\any}{dir.\xspace}
\newcommand{\anylong}{directed}

\newcommand{\undirected}{mutual\xspace}\newcommand{\Undirected}{\xmakefirstuc{\undirected}}
\newcommand{\topic}{q}
\newcommand{\user}{v}
\newcommand{\ofuser}{his/her\xspace}  
\newcommand{\auser}{he/she\xspace}  
\newcommand{\usersOnTopic}{\xmakefirstuc{\user}_\topic}
\newcommand{\tweets}{\mathit{tweets}} \newcommand{\tweetsi}{\tweets_{\user_i,\topic}}
\newcommand{\set}[1] {\{#1\}}
\newcommand{\setrestrict}[2]{\set{#1 \: | \: #2}}

\newcommand{\labelvar}{y}
\newcommand{\ulabel}[1]{\labelvar_{\user_#1}}
\newcommand{\vect}[1]{\mathbf{#1}}
\newcommand{\ulabels}{\vect{\xmakefirstuc{\labelvar}}}
\newcommand{\params}{\vect{\phi}}
\newcommand{\optfn}[1]{\Phi_\phi(#1)}
\newcommand{\partition}{Z}

\newcommand{\users}{V}
\newcommand{\nbrs}[1]{\mathit{Neighbors}_{\user_#1}}
\newcommand{\ione}{k}
\newcommand{\itwo}{\ell}
\newcommand{\indices}{\ione, \itwo}
\newcommand{\tlabelvar}{\hat{\labelvar}} 
\newcommand{\tlabel}[1]{\tlabelvar_{#1}}
\newcommand{\possent}{1} 
\newcommand{\negsent}{0}
\newcommand{\setsize}[1]{|#1|}

\newcommand{\wUnlabeled}{w_{{\rm unlabeled}}}
\newcommand{\wLabeled}{w_{{\rm labeled}}}
\newcommand{\wRelation}{w_{{\rm relation}}}
\newcommand{\utweight}[1]{\mu_{#1}} 
\newcommand{\uuweight}[1]{\lambda_{#1}} 
\newcommand{\utfactor}{f}\newcommand{\uufactor}{h}

\newcommand{\labelledEdges}{E_{\rm labeled}}
\newcommand{\sample}{{\sf Sample}}
\newcommand{\likeratio}{{\sf LLR}_\params}
\newcommand{\likeratioderiv}{\vect{{\nabla_\params}\likeratio}}
\newcommand{\relperf}{{\sf RelPerf}}
\newcommand{\oldlabels}{\ulabels}
\newcommand{\newlabels}{\ulabels^{\rm{new}}}
\newcommand{\learnRate}{\eta}
\newcommand{\perf}{{\sf Perf}}

\newcommand{\srank}{Learning\xspace} 
\newcommand{\sranklong}{SampleRank\xspace}
\newcommand{\sta}{NoLearning\xspace}
\newcommand{\stalong}{Direct estimation from simple statistics\xspace}
\newcommand{\labeled}{labeled\xspace}
\newcommand{\unlabeled}{unlabeled\xspace}

\newtheorem{definition}{Definition}
\newtheorem{problem}{Problem}

\title{
{
User-Level Sentiment Analysis Incorporating Social Networks
}
}

\numberofauthors{6}

\author{
{Chenhao Tan
}\\
Dept. of Computer Science\\
Cornell University\\
chenhao@cs.cornell.edu
\and
{Lillian Lee}\\
Dept. of Computer Science\\
Cornell University\\
llee@cs.cornell.edu
\and 
{Jie Tang}\\
Dept. of Computer Science\\
Tsinghua University\\
jietang@tsinghua.edu.cn
\and
{Long Jiang}\\
Microsoft Research Asia\\
Microsoft Corporation\\
pkujianglong@hotmail.com
\and
{Ming Zhou}\\
Microsoft Research Asia\\
Microsoft Corporation\\
mingzhou@microsoft.com
\and 
{Ping Li}\\
Dept. of Statistical Science\\
Cornell University\\
pingli@cornell.edu
}

\maketitle
\sloppy
\vfil\eject

\begin{abstract}
  We show that information about social relationships can be used to
  improve user-level sentiment analysis. 
The main motivation behind our approach is that users that are somehow
``connected'' may be more likely to hold similar opinions; therefore,
relationship information can complement what we can extract about a user's
viewpoints from their utterances.
  Employing Twitter as a source for our experimental data,
and working within a semi-supervised framework, 
 we propose
  models that are induced either
from the Twitter follower/followee
  network or 
from the 
network in Twitter formed by users referring to
  each other using ``@'' mentions.
 Our
transductive learning
  results reveal that incorporating social-network information can
  indeed lead to statistically significant sentiment-classification
  improvements over the performance of 
an approach based on 
Support Vector Machines having
  access only to textual features.
\end{abstract}




\section{Introduction}
\label{sec:intro}

Sentiment analysis \cite{Pang+Lee:08b} is one of the key emerging
technologies in the effort to help people navigate the huge amount of 
user-generated content available online. 
Systems that automatically determine viewpoint would
enable users to
 make sense of the
enormous body of opinions expressed on the Internet, 
ranging
from product
reviews to political
positions.

We propose to
improve sentiment analysis by utilizing the
information about 
user-user relationships made evident by online social networks.
We do so for two reasons.  First, 
user-%
relationship information is now more easily
obtainable, since  user-generated content often appears in the context
of social media. For example, Twitter maintains not just tweets, but
also lists of followers and followees. 
Second, and more importantly, 
when a user forms a link in a network such as Twitter, they do so to create a connection.  If this connection corresponds to a personal relationship, then
the
 principle of \emph{homophily} \cite{Lazarsfeld+Merton:1954a}
--- 
the idea that similarity and  connection tend to co-occur, or
 ``birds of a feather flock together''
 \cite{McPherson+Smith-Lovin+Cook:2001a} --- suggests that users that
 are ``connected'' 
by a mutual personal relationship
may tend to hold similar opinions;
indeed, one
 study found some evidence of homophily for both positive and negative
 sentiment among MySpace Friends \cite{Thelwall:2010a}.  
Alternatively, the connection a user creates may correspond to approval (e.g., of a famous figure) or a desire to pay attention (e.g., to a news source), rather than necessarily a personal relationship; but such connections are still also suggestive of the possibility of a shared opinion.

Therefore, employing Twitter as the basis for our sentiment classification experiments, we incorporate
user-%
relation
 information, as follows.  We first 
utilize a
model based on the follower/followee network that has dependencies not
only between the 
opinion of  
a user and the opinions
expressed in
\ofuser tweets, 
but also between 
\ofuser opinion
and 
those of the users that 
\auser follows.
 We also
consider an @-network-based variant, in which we have dependencies between 
a user's opinion
 and the opinions of those whom 
\auser 
mentions via
an ``@''-reference.

We work within a semi-supervised, user-level framework.  The reason we adopt a
semi-supervised approach is that  the acquisition of a large quantity
of relevant sentiment-labeled data can be a time-consuming and
error-prone process, as discussed later in this paper. We focus on
user-level rather than tweet-level (corresponding to document- or
sentence-level) sentiment because the end goal for many users of
opinion-mining technologies is to find out what \emph{people} think;
determining the sentiment expressed in individual texts is usually a subtask of
or proxy for that ultimate objective. Additionally, it is plausible
that there are cases where some of a user's tweets are genuinely
ambiguous (perhaps because they are very short),
but 
\ofuser overall opinion can be determined by looking at
\ofuser collection of tweets and 
who 
\auser is connected to.

\vpara{Contributions} %
First, we empirically confirm 
that the probability that two users
share the same opinion  is indeed correlated
with whether they are connected in the social network. 

We then show that using graphical models incorporating social-network
information can lead to statistically significant improvements in
user-level sentiment polarity classification with respect to 
an approach
 using only textual information.

Additionally, we perform an
array of experimental comparisons
that  encompasses not only the variation in
underlying 
network (follower/followee vs.~@-network) mentioned above,
but also variation in 
how the parameters of our model are learned; how the
baseline SVMs are trained; and 
which graph we employ, 
i.e., is it enough for user $\user_i$ to follow user $\user_j$
 (corresponding to attention and/or homophily),
or should
we require that $\user_i$ and $\user_j$ 
mutually follow each
other
 (more in line with homophily only)?
For some topics,  a combination of homophily and approval/attention
links 
performs better
than homophily links alone; but in other topics, homophily-only links are best. Interestingly, we find that when the edge quality is sufficiently high, even a very small number of edges can lead to statistically
significant improvements.

\vpara{Paper organization} 
\S \ref{sec:problem} gives a formal characterization of our user-level sentiment categorization problem
within the setting of Twitter.
\S \ref{sec:observation} introduces the data set we collect and
presents 
some motivational analysis.
\S\ref{sec:approach} explains
our proposed model and
describes
 algorithms for parameter estimation and prediction. 
\S \ref{sec:exp} presents 
our experimental results.
\S \ref{sec:related} introduces related work not mentioned otherwise. 
\S \ref{sec:conclude} concludes by summarizing our work and discussing future directions.

\section{Concrete Problem Setting}
\label{sec:problem}
In this section, we frame the problem
in the context of Twitter to keep things concrete,
 although adaptation of this framework to
other social-network settings is straightforward.   
In brief, we address the semi-supervised topic-dependent
sentiment-polarity user categorization task.
In
doing so, we consider
four different ways in which Twitter users can be considered to be ``connected''.

Our task is to classify each user's sentiment on a specific topic into
one of two polarities:
``Positive'' and 
``Negative''.\footnote{We initially worked with
  positive/negative/neutral labels, but determining neutrality was
  difficult for the annotators.} ``Positive'' 
means that the user supports or likes the target
topic, 
whereas
``Negative'' stands for the opposite.
(As stated above, this differs from classifying  each of a user's tweets.)
Given the scale of  Twitter and the difficulty in acquiring labels (see \S \ref{sec:observation}), we work within the semi-supervised learning paradigm.
That is, we assume that we are given a topic and a user graph, where a
relatively small proportion of the users have already been labeled,
and the task
is to predict the labels of all the unlabeled users.

Our motivating intuition, that ``connected'' users will tend to hold similar opinions, requires us to define what ``connected'' means.  For Twitter, there are several possibilities.
These roughly correspond to whether we should consider only ``personal connections'', in accordance with homophily, or ``any connection'', which is more in line with the approval/attention hypothesis mentioned in the introduction.  Note that focusing on personal connections presumably means working with less data.

The first possibility we consider is to deem two
Twitter
 users to be connected if one 
``follows'' the other.  (From now on, to distinguish between the Twitter-defined ``following'' relationship and ordinary English usages of the word ``follow'', we use 
\emph{``\follow''} to refer to the Twitter version.)  
This corresponds to the idea 
that  users often agree with those they pay attention to.  Of course, this isn't always true: for example,  21\% of US Internet users usually consult websites that hold opposing political viewpoints \cite{Smith:2010a}.
So, alternatively, 
we may instead 
 only consider pairs of users who know each other
personally.  
As a rough proxy for this sort of relationship information, we look at whether a user mentions another by name using the Twitter @-convention; the intuition is that a user will address those who they are having a conversation with,
and thus know. 
Again, though, this is only a heuristic.

Another factor to take into account is whether we should require both users in a potential pair to connect with each other. 
Mutual connections presumably indicate stronger relationships,
but attention effects may be more important than homophily effects with respect to shared sentiment.

We thus have  2 $\times$ 2 possibilities for our definition of when
we decide that a connection (edge) between users exists.

\begin{itemize}
\setlength{\itemsep}{-0.2\baselineskip}
\item \emph{Directed \follow graph}: user $\user_i$ \follows $\user_j$
($\user_j$ may or may not \follow $\user_i$ in return).
\item \emph{\Undirected
\follow graph}: user $\user_i$ \follows $\user_j$
\emph{and} user $\user_j$ \follows $\user_i$.
\item \emph{Directed @ graph}: $\user_i$ has mentioned $\user_j$ via a tweet containing
  ``@$\user_j$''  
($\user_j$ may or may not @-mention $\user_i$ in return).
\item  \emph{\Undirected 
@ graph}: $\user_i$ has mentioned $\user_j$ via a tweet containing
  ``@$\user_j$'' \emph{and} vice versa.
\end{itemize}

\section{Data and Initial Observations}
\label{sec:observation}

\subsection{Data Collection}

\vpara{Motivation}
We first planned to adopt the straightforward approach to creating a
\labeled test set, namely, manually annotating arbitrary Twitter users
as to their sentiment on a topic
by reading
the users' on-topic
 tweets.  However, inter-annotator agreement was
far below what we considered to be usable levels.  Contributing factors
included the need for familiarity with
topic-specific information and 
 cultural context
to interpret individual tweets;
for example, 
the tweet ``\#lakers
 b**tch!''  was
  mistakenly labeled negative for the topic 
``Lakers''
(the expletive was spelled out in full in the original).

Fortunately,
this problem can be
avoided to some degree
 by 
taking advantage of the fact
that user metadata is often
much easier to interpret. 
For example, with respect to the topic ``Obama'', there are  Twitter
users with the bios ``social engineer, karma dealer, \& obama lover'' and
``I am a right wing radical-American that is anti-Sharia law,
anti-muslim, pro-Israel, anti-Obama and America 
FIRST'',
another with username ``against\_obama'', and so
on.\footnote{In practice, we also require that such
  users actually tweet on the topic.}
  We thus  employed the following data-acquisition procedure.

\vpara{Initial pass over users}
Our goal was to find a large
set of users whose opinions are clear, so that the gold-standard
labels
would be reliable.
To begin the collection process, we
selected
as seed users 
 a set of
high-profile
political figures and a set of users
who seemed opposed to them
(e.g., ``BarackObama'',  ``RepRonPaul'', ``against\_obama''). 
We performed a crawl by traversing edges starting from our seed set.

\vpara{Topic selection and 
gold-standard
user labeling
} 
In the crawl
just described, 
the set of profiles containing the corresponding keyword tended to be
hugely biased towards the positive class.
We therefore used the initially-gathered profiles to try to find
topics with a more balanced class distribution:
we computed those keywords with the
highest frequencies among the words in the profiles,  resulting in the
topics ``Obama'', ``Sarah Palin'', ``Glenn Beck'',
``Fox News'', and ``Lakers''
 (e.g., 
``Ron Paul'' was not in this final set). 
Then we 
employed a very conservative 
strategy:
we annotated
 each user according to their biographical
information
(this information was \emph{not} used in our algorithms),
keeping
only
 those whose opinions we could clearly determine  from their name and 
bio.
\footnote{
When the strictness of these constraints led to a small result set for some
topics, we augmented the \labeled dataset with more users
whose labels 
could be 
determined by examination of their tweets.  In the case of ``Lakers'', we were able to acquire more negative users by treating users with  positive sentiment towards ``Celtics'' as negative for
  ``Lakers'', since the Celtics and Lakers are 
two traditional rivals among US basketball teams.}
This approach does mean that 
we are working with graphs in which the users have
strong opinions on the target topic, but 
the resulting gold-standard sentiment labels will be trustworthy.

\vpara{
Resultant 
graphs}
Finally, we constructed the graphs for our main experiments from the users with gold-standard labels and the edges between them.
Table \ref{tb:basicstatistics} shows basic statistics across topics.
 ``On-topic
tweets'' 
means
 tweets
mentioning the topic by the name we assigned it (e.g., a tweet
mentioning ``Barack'' but not  ``Obama'' would not be included): our
experiments only consider on-topic tweets.

\begin{table}[tb!]
\centering \caption{\label{tb:basicstatistics} 
Statistics for our main datasets.
}
\scriptsize
\begin{tabular}{|c||r||r|r|r|r||r|}
\hline 
{Topic} & \# users & \multicolumn{2}{c|}{\#\follow edges} &  \multicolumn{2}{c||}{\#@ edges} & \#  on-topic tweets\\
    \cline{3-6}
                      &                           & \multicolumn{1}{c|}{\any} & \undirected   & \multicolumn{1}{c|}{\any} & \undirected  & \\ 
\hline \hline Obama & 889 & 7,838 & 2,949 & 2,358 & 302 & 128,373\\
\hline Sarah Palin & 310 & 1,003 &264 & 449 & 60 & 21,571\\
\hline Glenn Beck &  313 & 486 & 159 & 148 & 17 & 12,842\\
\hline Lakers & 640 & 2,297 & 353 & 1,167 & 127 & 35.250\\
\hline Fox News & 231 & 130 &32 & 37 & 5 & 8,479\\
\hline
\end{tabular}
\normalsize
\end{table}

\subsection{Observations}

Before proceeding, we first engage in some high-level investigation
of 
the degree to which
network structure and user labels correlate, 
since a major motivation for our work is the intuition that connected
users tend to exhibit similar sentiment.
We study 
the interplay between
user labels and network influence via the following two kinds of statistics:
\begin{enumerate}
\setlength{\itemsep}{-0.2\baselineskip}
\item Probability that two users have the same label, conditioned
  on whether or not they are connected
\item 
Probability that two users are connected, 
conditioned on whether or not  they have the same label
\end{enumerate}
 As 
stated
 in \S 
\ref{sec:problem}, 
we have four types of user-user connections to
consider: 
\follow and mutually-\follow  relationships, and
@-mentioned and mutually-@-mentioned relationships. 

\begin{figure}[htb!]
\begin{center}
\begin{tabular}{cc}
\Follow Graph & @ Graph\\
\epsfig{figure=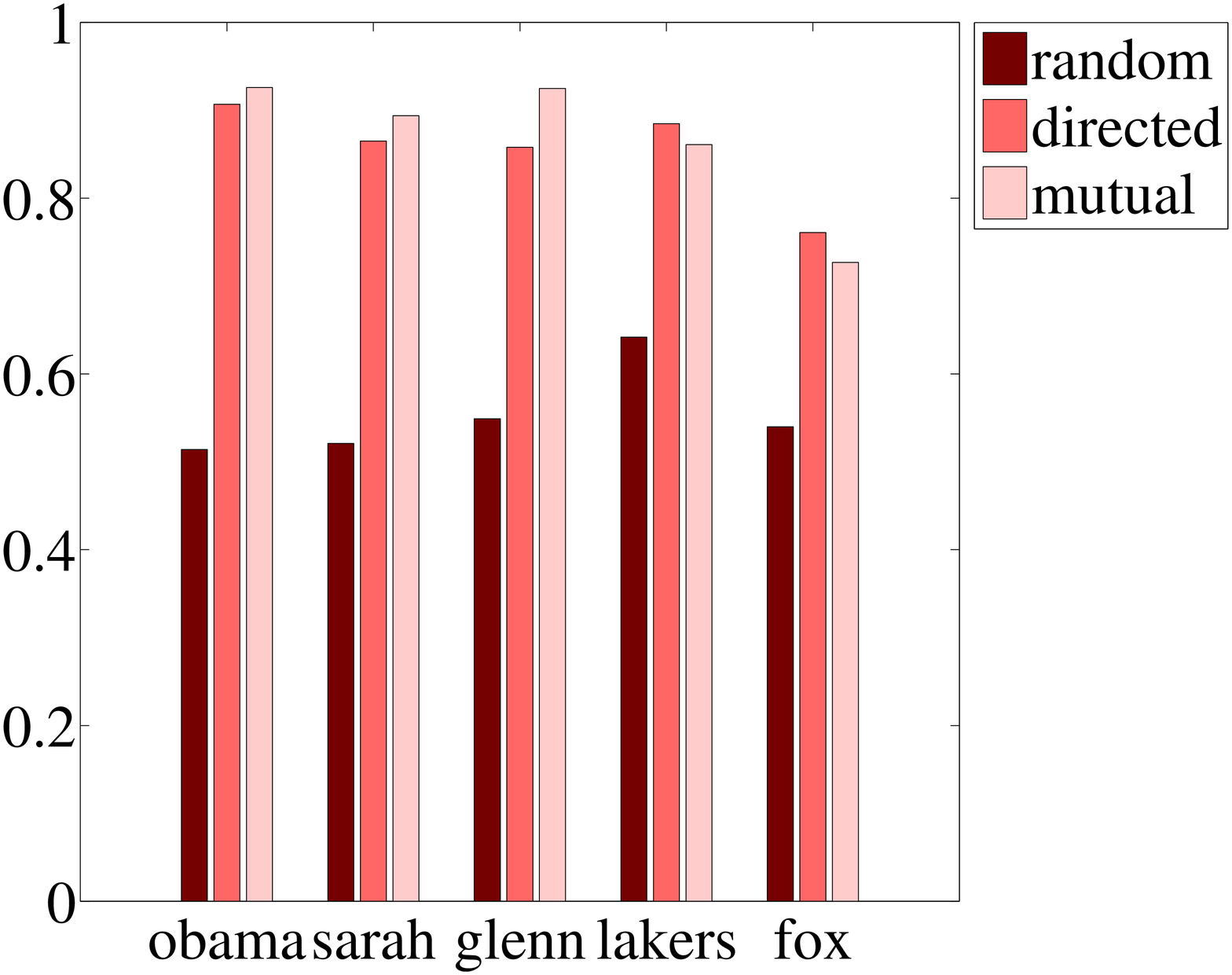, width=1.5in} &
\epsfig{figure=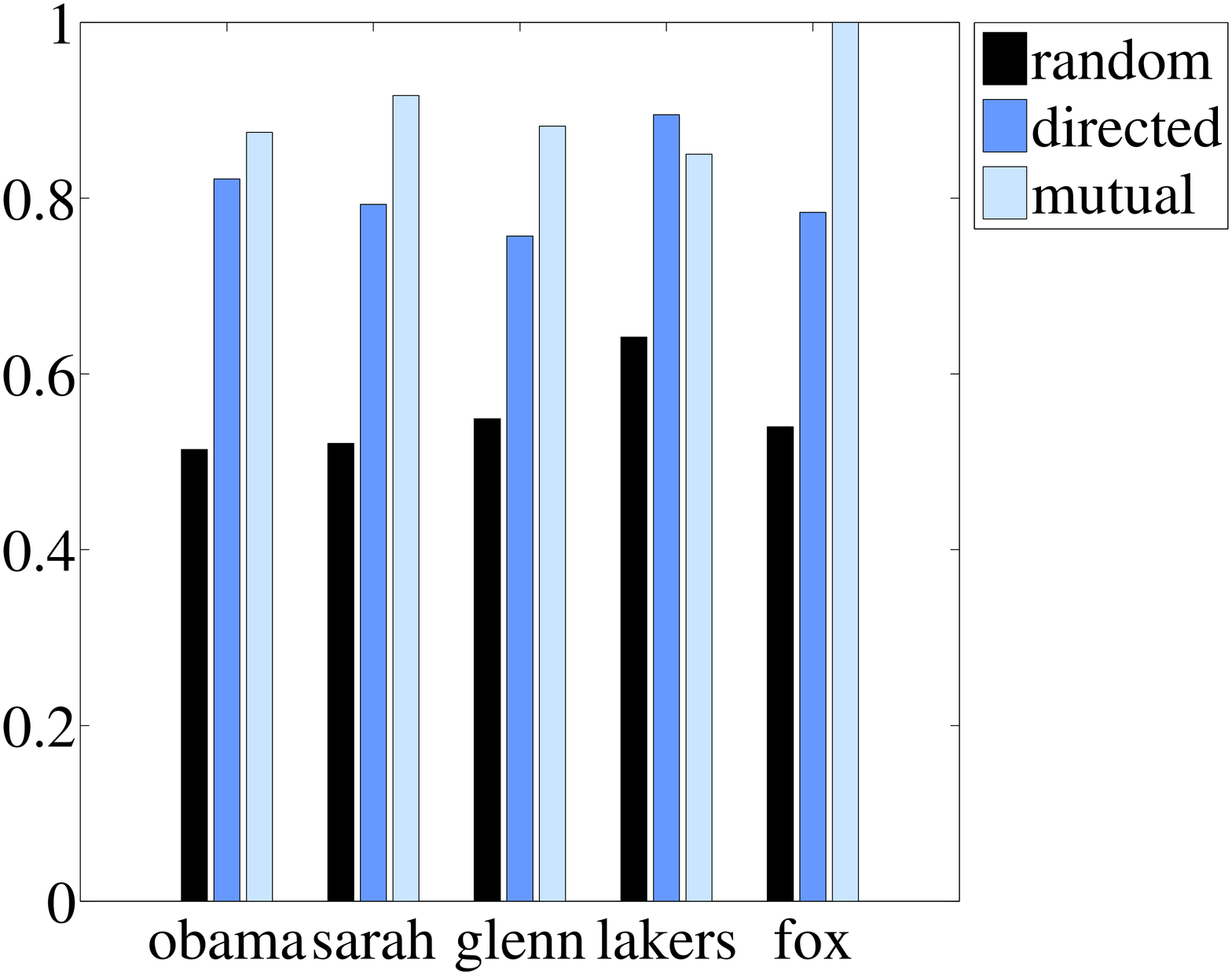, width=1.5in}\\
\end{tabular}
\end{center}
\caption{\label{fig:sentimentStatistics}
Shared sentiment conditioned on type of connection.
\small Y-axis: probability of two users
$\user_i$ and $\user_j$
having the same sentiment label, conditioned on 
relationship type.  
The left
plot
is for 
the \follow graph, while the right one is for the @ graph. 
``random'': 
pairs formed by randomly choosing users.
``directed'': 
at least one user in the pair links to the other.   
``\undirected'': 
both users in the pair link to each other.
Note that the very last bar (a value of 1 for ``Fox News'', \undirected @-graph)
is based on only 5 edges (datapoints).
}
\normalsize
\end{figure}

\vpara{
Shared sentiment
conditioned on being connected}
Figure \ref{fig:sentimentStatistics} 
clearly shows that the probability of two connected users sharing the
same sentiment on a topic is much higher than chance. 
The effect is 
a bit
more pronounced  overall in the 
\follow graph (red
bars) than in the @-graph (blue bars): for instance, more of the
bars are greater than .8. 
In terms of ``\undirected'' links
(mutual \follow or @-mentions)
compared with ``directed'' links,
where the \follow or @-mentioning 
need not be
mutual, 
it is interesting to note that ``\undirected'' 
corresponds to a  higher probability of shared sentiment in the topics
``Obama'', ``Sarah Palin'',  and ``Glenn Beck'',
while the reverse holds for 
``Lakers''.

\begin{figure}[htb!]
\begin{center}
\begin{tabular}{cc}
\Follow Graph & @ Graph\\
\epsfig{figure=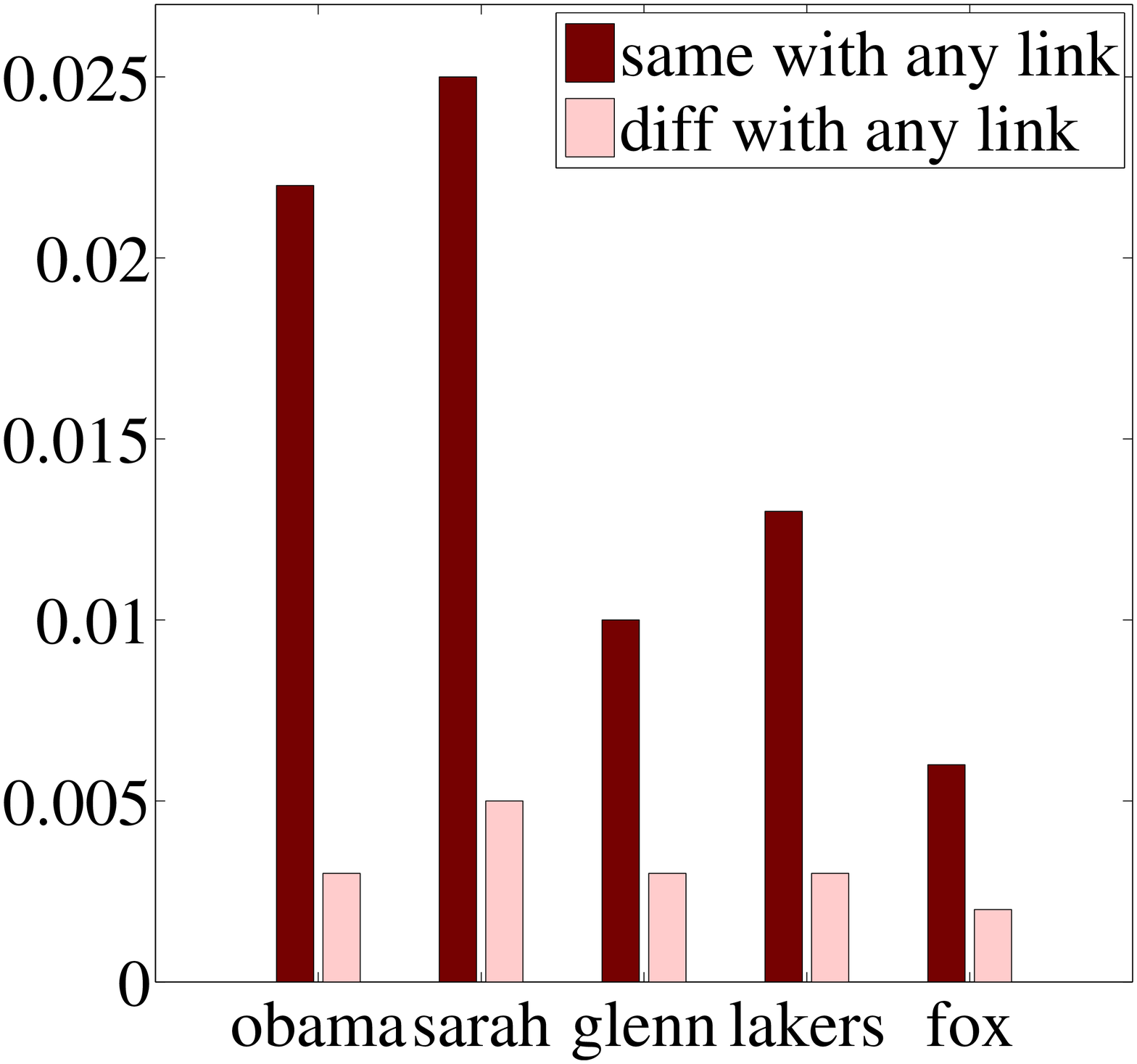, width=1.5in} &
\epsfig{figure=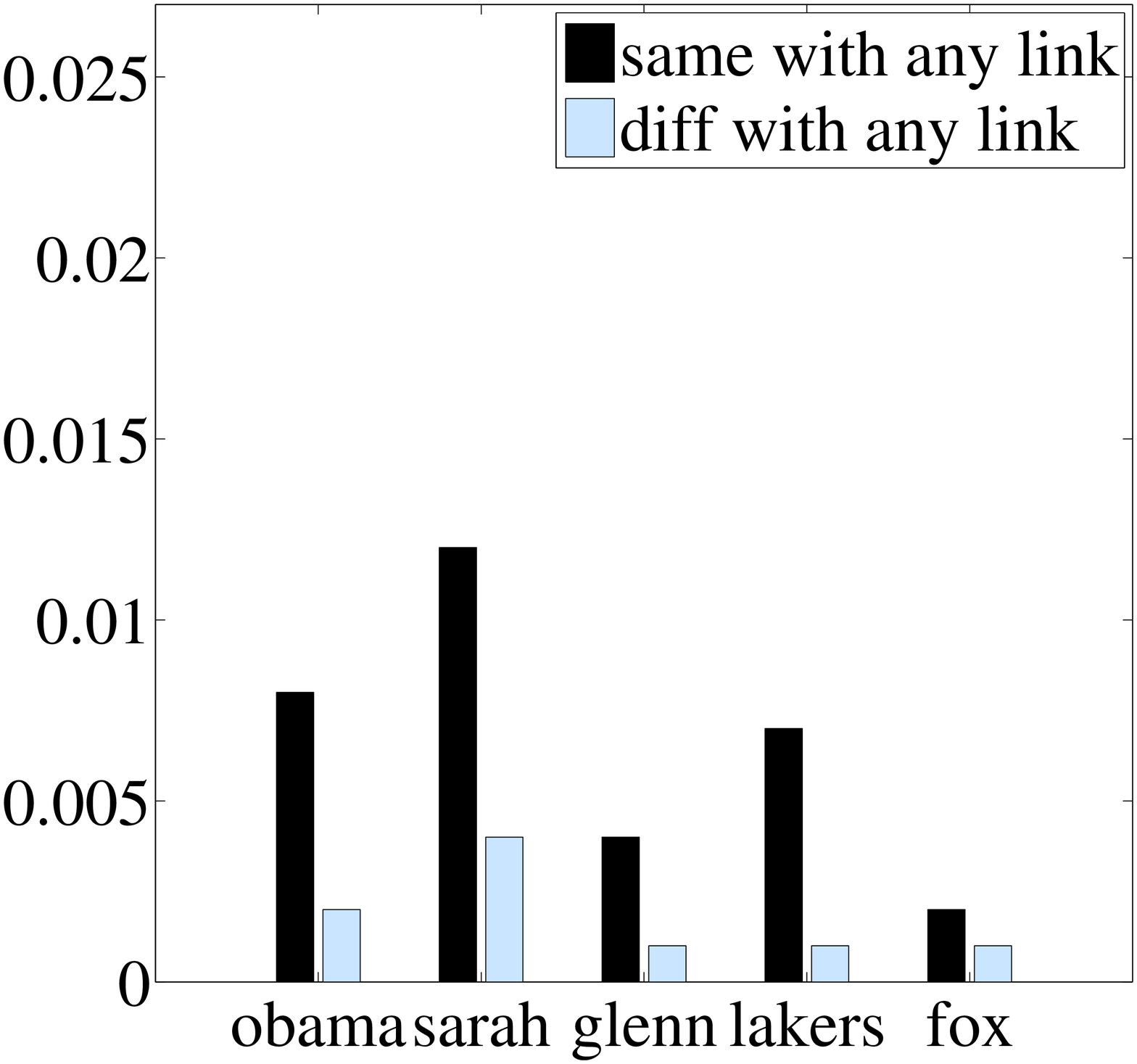, width=1.5in}\\
\end{tabular}
\end{center}
\caption{\label{fig:linkStatistics}
Connectedness conditioned on labels.
 \small Y-axis: probability that two users are
connected, conditioned on whether or not the users have the
same sentiment. 
\vspace{-.2in}
}
\normalsize
\end{figure}

\vpara{Connectedness conditioned on labels}
We now turn to our second statistic, which measures whether shared
sentiment tends to imply connectedness.  Figure
\ref{fig:linkStatistics} clearly shows
that in our graphs, 
it is much more likely for
users to be connected if they share an opinion than if they differ.
The probability that same-opinion users are connected is much larger in the 
\follow
graph than in the @ graph.
This may be a result of the fact that the @-graph is more sparse, as can be seen from Table \ref{tb:basicstatistics}.

\vpara{Summary} 
We have seen 
that first,
user pairs in which at least one party links to the other
are more likely to hold the
 same sentiment,
and second, 
two users with the same sentiment are more 
likely
to have at least one link to the other
 than
two users with different sentiment. These points validate 
our intuitions that 
links and shared sentiment are clearly correlated, at least in our data.

\vspace*{-.1in}

\section{Model Framework}
\label{sec:approach}

In this section, we 
give a formal definition of
the 
model we work with.
We propose a 
factor-graph model for user labels.
With our formulation,
more-or-less standard technologies can be employed for learning and
inference. %
We employ transductive learning algorithms in our models.
The main advantage of our formulation is that it employs social-network
structure to help us overcome
both the paucity of textual information in short tweets and the lack of a large
amount of \labeled data. 

\vspace*{-.05in}

\subsection{Formulation}

We are given a ``query'' topic $\topic$, 
which induces a set of users $\usersOnTopic$ who have tweeted about $\topic$.\footnote{We omit
  users who have never expressed an opinion about  $\topic$; it seems
  rash to judge someone's  opinion based {\em solely} on who their
  associates are.} Our goal is to determine which users in $\usersOnTopic$
are positive towards $\topic$ and which are negative.

 For each user $\user_i \in
\usersOnTopic$, we have the set $\tweetsi$ of $\user_i$'s tweets
about $\topic$, and we know which users 
$\user_j 
\in \usersOnTopic$
\follow 
or @-mention
$\user_i$ and vice versa. 
Recall that we are working in a semi-supervised setting where
we are given sentiment labels on 
a relatively small subset
 of the users in
$\usersOnTopic$.  (We do not assume sentiment labels on the tweets.)

We incorporate both textual and 
social-network information in a
single {\em heterogeneous graph on topic $q$}, where nodes can
  correspond to either users or 
tweets.   Figure \ref{fig:heterogeneousExample} shows an
example. 
\begin{definition}
A \textbf{heterogeneous graph on topic $q$} is a graph
$HG_\topic=\set{\usersOnTopic \cup 
\setrestrict{\tweetsi}{\user_i \in \usersOnTopic}, 
E_\topic}$.   The edge set $E_q$ is the union of two sets: the tweet edges
\mbox{$\setrestrict{(\user_i, \tweetsi)}{\user_i \in \usersOnTopic}$},
indicating that $v_i$ 
posted
$\tweetsi$,
and 
network-induced 
user-user
edges. 
\end{definition}
As already mentioned in \S \ref{sec:problem}, 
we consider four types of heterogeneous graphs, characterized by the
definition of 
when socially-induced edge $(v_i,
 v_j)$
exists in $E_\topic$:
{directed \follow}, {\undirected \follow}, {directed @}, and
{\undirected @
graphs.

Tweet edges are taken to 
be
either directed or undirected to match the
type of the socially-induced edges.

\begin{figure}
\begin{center}
\input{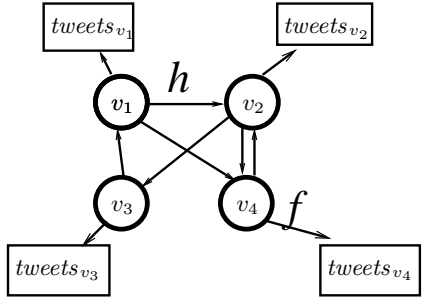}
\end{center}
\caption{\label{fig:heterogeneousExample} 
Example directed heterogeneous graph (dependence on topic $\topic$ suppressed
for clarity). 
\small The corresponding factor graph has factors corresponding to user-tweet
dependencies (label ``f'') and user-user dependencies (label ``h'').
}
\end{figure}
 
\subsection{Proposed Model}

Let the topic be fixed, so that we can suppress it in the notation
that follows and say that we are working with heterogeneous graph $HG$
involving a set of users $\users = \set{\user_i}$. Let $\ulabel{i}$ be
the label for user $\user_i$, and let
\renewcommand{\tweetsi}{\tweets_{\user_i}}
$\ulabels$ be the vector of labels for all users.
We make the Markov assumption
that the user sentiment $\ulabel{i}$ is influenced only by the
(unknown) sentiment labels $\tlabel{t}$ of  tweets $t \in \tweetsi$ and the
(probably unknown) sentiment labels of the immediate 
user
neighbors
$\nbrs{i}$ of $\user_i$. This assumption leads us to the following
factor-graph-based model:
\begin{equation}
\label{eq:userProb}
\begin{split}
\log P(\ulabels)  =   
\big(\sum_{\user_i \in \users}
 \big[&\sum_{t \in \tweetsi,
  \indices} \utweight{\indices}
 \utfactor_{\indices}(\ulabel{i}, \tlabel{t})  
\\ 
 & + \sum_{\user_j \in \nbrs{i}, \indices} \uuweight{\indices} 
\uufactor_{\indices}(\ulabel{i}, \ulabel{j})\big] \big) \\
- 
\log{\partition}
,
\end{split}
\end{equation}
where the first and second 
inner sums
correspond to user-tweet factors and
user-user factors, respectively  (see below for more details), and the indices $\indices$ range over
the set of sentiment labels \set{\negsent,\possent}.
  $\utfactor_{\indices}(\cdot, \cdot)$ and $\uufactor_{\indices}(\cdot, \cdot)$
are feature functions, and $\utweight{\indices}$ and  $\uuweight{\indices}$ are
parameters representing impact.  (For instance, we might set $\utweight{0,1}$
to 0 to 
give no credit to 
 cases in which user label 
$\ulabel{i}$ is 0 but tweet
$t$'s label $\tlabel{t}$ is 1.) 
 $\partition$ is the normalization factor.

\vpara{User-tweet factor} 
Feature function
$\utfactor_{\indices}(\ulabel{i},\tlabel{t})$ 
fires for
a particular configuration, 
specified by the indices $\ione$ and $\itwo$,  of user
and individual-tweet labels 
(example configuration:
both are \possent). 
After all, we expect $\user_i$'s  tweets to provide information about
their opinion.
 Given our semi-supervised setting, we
opt to give different values to the same configuration depending on
whether
or not
 user $\user_i$ was one of the initially \labeled items,
the reason being that 
the initial labels are probably more dependable.
Thus, we use $\wLabeled$ and $\wUnlabeled$ to indicate our different levels of confidence in users that were or were not initially labeled:
 \begin{equation}\label{Eq:userTweet}
\utfactor_{\indices}(\ulabel{i},\tlabel{t})  = \left\{
{\begin{array}{ll}
   {\frac{\wLabeled}{\setsize{\tweetsi}}} & {\ulabel{i} = \ione, \tlabel{t}=\itwo, \user_i~\mbox{labeled}}  \\
   {\frac{\wUnlabeled}{\setsize{\tweetsi}}} &  {\ulabel{i} = \ione,
     \tlabel{t}=\itwo, \user_i~ \mbox{unlabeled}}  \\
   {0} & \mbox{otherwise} \\
\end{array} } \right.
\end{equation}
We normalize by $\setsize{\tweetsi}$ because each $t \in \tweetsi$
contributes to the first exponential in Eq. \ref{eq:userProb}.

\vpara{User-user factor}
Next, our observations in \S \ref{sec:observation} suggest that 
social-network
connections between users can correlate with agreement in sentiment.
Hence, we define feature functions 
 $\uufactor_{\indices}(\ulabel{i},
\ulabel{j})$, which  
fire for
a particular configuration of labels, specified by the indices $\ione$ and
$\itwo$,  between neighboring users $\user_i$ and $\user_j$:
\begin{equation}\label{Eq:userUser}
\uufactor_{\indices}(\ulabel{i},\ulabel{j})  = \left\{
{\begin{array}{ll}
   {\frac{\wRelation}{\setsize{\nbrs{i}}}} & {\ulabel{i} = \ione, \ulabel{j}=\itwo} \\
   {0} & \mbox{otherwise} \\
\end{array} } \right.
\end{equation}

Note that for a directed heterogeneous graph with edge set $E$, we define $\nbrs{i} \stackrel{def}{=}
\setrestrict{\user_j}{(\user_i,\user_j) \in E}$, since the Twitter
interface makes the tweets of {\follow}ee $\user_j$ visible to {\follow}er $\user_i$
(and similarly for @-mentions), so we have some reason to believe that
$\user_i$ is aware of $\user_j$'s opinions.

\vpara{Implementation Note}
in our experiments, we empirically set the
weights within the feature functions as follows:  $\wLabeled=1.0$,
$\wUnlabeled=0.125$, 
$\wRelation=0.6$;\footnote{
These parameters are set to adjust the importance of labeled data, unlabeled data and relation information. 
We
did try
different 
parameter values.
In accordance with the
 intuition that labeled users are the most
trustworthy,
and that user relations are the next most trustworthy, 
we 
fixed
$\wLabeled=1.0$, 
and then varied
$\wRelation$ between 
$[0.5, 0.8]$ and  $\wUnlabeled$ 
between $[0.1, 0.5]$. 
The parameter settings given in the main text exhibited
the best performance across topics, but performance was relatively
stable
across different settings.}
thus, the greatest emphasis is
on tweet labels matching the label of an initially-labeled user.

\subsection{Parameter Estimation and Prediction}
\label{sec:alg}

We now address the problems of estimating the remaining
free parameters and inferring user sentiment labels once the parameter
values have been learned.  We provide more details below, but to
summarize: Inference is performed using loopy belief
propagation, and  for parameter estimation, we employ two
approaches.  The first is simple estimation from the small set of \labeled
data we have access to; the second applies 
\sranklong to the semi-supervised setting \cite{wick09sample,Singh+Yao+Riedel+McCallum:10}.

\subsubsection{Parameter Estimation}

To avoid needing to
always
 distinguish between $\utweight{\indices}$'s and $\uuweight{\indices}$'s,
 we introduce a change of notation. 
We write $\params$ for the vector of 
 parameters
$\utweight{\indices}$ and $\uuweight{\indices}$, and write
$\optfn{\ulabels}$ for the function $\log P(\ulabels)$,
given in Eq.\ref{eq:userProb}, 
induced by a particular $\params$ on a vector of user labels
$\ulabels$.    If we were in the fully supervised setting --- that is, if
we were given $\ulabels$ ---  then in
principle all we would need to do is find the $\params$ maximizing
$\optfn{\ulabels}$; but recall that we are working in a
semi-supervised setting. 
We propose the following two approaches.

\vpara{\stalong (``\sta'' for short)} %
One way around this problem is to not learn the parameters $\params$ via
optimization, but to  simply use counts from the \labeled subset of the data. 
Thus, letting $\labelledEdges$ be the subset of edges in our
heterogeneous graph in which both endpoints are \labeled, we estimate the four
user-user 
parameters as follows:
\begin{equation}
\scriptsize
\uuweight{\indices} := \frac{\sum_{(\user_i,\user_j) \in \labelledEdges} I(\ulabel{i}=\ione, \ulabel{j}=\itwo)}{\sum_{(\user_i,\user_j) \in \labelledEdges} I(\ulabel{i}=\ione, \ulabel{j}=\possent) + I(\ulabel{i}=\ione, \ulabel{j}=\negsent)} 
\end{equation}
where $I(\cdot)$ is the indicator function.
Remember, though, that we do not have any labels on  the (short,
often hard-to-interpret) tweets.  We therefore make the 
strong
assumption that positive users only post positive (on-topic) 
tweets,
and negative users only post negative tweets; we thus set
$\utweight{\indices} := 1$ if $\ione = \itwo$, 0 otherwise.

\vpara{\sranklong (``\srank'')} 
If we instead seek to learn the parameters $\params$ by maximizing $\optfn{\cdot}$, we need to
determine how to deal with the normalization factor and how to best handle
having both \labeled and \unlabeled data.  
We employ \sranklong \cite{wick09sample},  Algorithm \ref{alg:em}:
\begin{algorithm} [h]
\label{alg:em}
\caption{
\sranklong. 
In our experiments, $\eta$= .001; varying $\eta$ did not affect
performance much.
}
\scriptsize
\SetLine 
\KwIn{Heterogeneous graph $HG$ with labels on some of the user nodes,
  learning rate $\learnRate$}
\KwOut{Parameter values $\params$  and full label-vector $\ulabels$}\BlankLine

Randomly initialize $\oldlabels$;

Initialize 
$\params$ from \sta;

  \For{$i := 1$ to \emph{Number of Steps}}
  {
    $\newlabels := \sample(\oldlabels)$\;

    \If{($\relperf(\newlabels, \oldlabels) > 0$ and   $\likeratio(\newlabels, \oldlabels) < 0$) \newline
//performance is better 
but the objective function is lower
\newline
or ($\relperf(\newlabels, \oldlabels) < 0$ and   $\likeratio(\newlabels, \oldlabels) > 0$)\newline
//performance is worse 
but the  objective function is higher\newline
    }
    {
       $\params :=  \params  - \learnRate  \likeratioderiv(\newlabels, \oldlabels)$;
    }
    \If{convergence}
    {
        break;
    }
    \If{$\relperf(\newlabels, \oldlabels) > 0$} 
    {
        $\ulabels:=\newlabels$\;
    }
  }
\normalsize
\end{algorithm}

\noindent In the above, 
$\sample$ is the sampling function; we use
  the uniform distribution in our experiments 
.
$\likeratio(\newlabels, \oldlabels)$ is the log-likelihood
ratio for the new sample $\newlabels$ and previous label set $\oldlabels$:
$\likeratio(\newlabels, \oldlabels) =
\log\left(\frac{P(\newlabels)}{P(\oldlabels)}\right)  = \optfn{\newlabels} - \optfn{\oldlabels}$
(this causes the 
normalization
terms
to
 cancel out). 
We can use all the users, \labeled and \unlabeled,  to compute
$\likeratio(\newlabels,\oldlabels)$, since we only need the underlying
graph structure to do so (the label sets to be compared are inputs to
the function).

We define the relative-performance or truth function
$\relperf(\newlabels,
  \oldlabels)$ as the difference in performance, measured on the
  {\em \labeled data only}, between $\newlabels$ and $\oldlabels$, where
the performance $\perf$ of a set of labels $\ulabels$ is
$\perf(\ulabels) = {\sf Accuracy}_{\rm \labeled}(\ulabels) + {\sf
  MacroF1}_{\rm \labeled}(\ulabels)$.
\citet{Singh+Yao+Riedel+McCallum:10} propose a  more sophisticated
approach to defining truth functions in the semi-supervised setting,
but our emphasis in this paper is on showing that our model is
effective even when deployed with simple learning techniques.

\subsubsection{Prediction}

We employ loopy belief propagation to perform inference for a given
learned model\footnote{
Using SampleRank for inference led to worse performance.}, as handling the normalization factor $\partition$ is
still difficult.  To 
account
 for the fact that
\sranklong
is randomized, we do  learning-then-inference 5
times to get 5 predictions, and take a majority vote among the five
label possibilities for each user.

\section{Experiments}
\label{sec:exp}

\begin{figure*}[htb!]
\begin{center}
\begin{tabular}{ccc}
\epsfig{figure=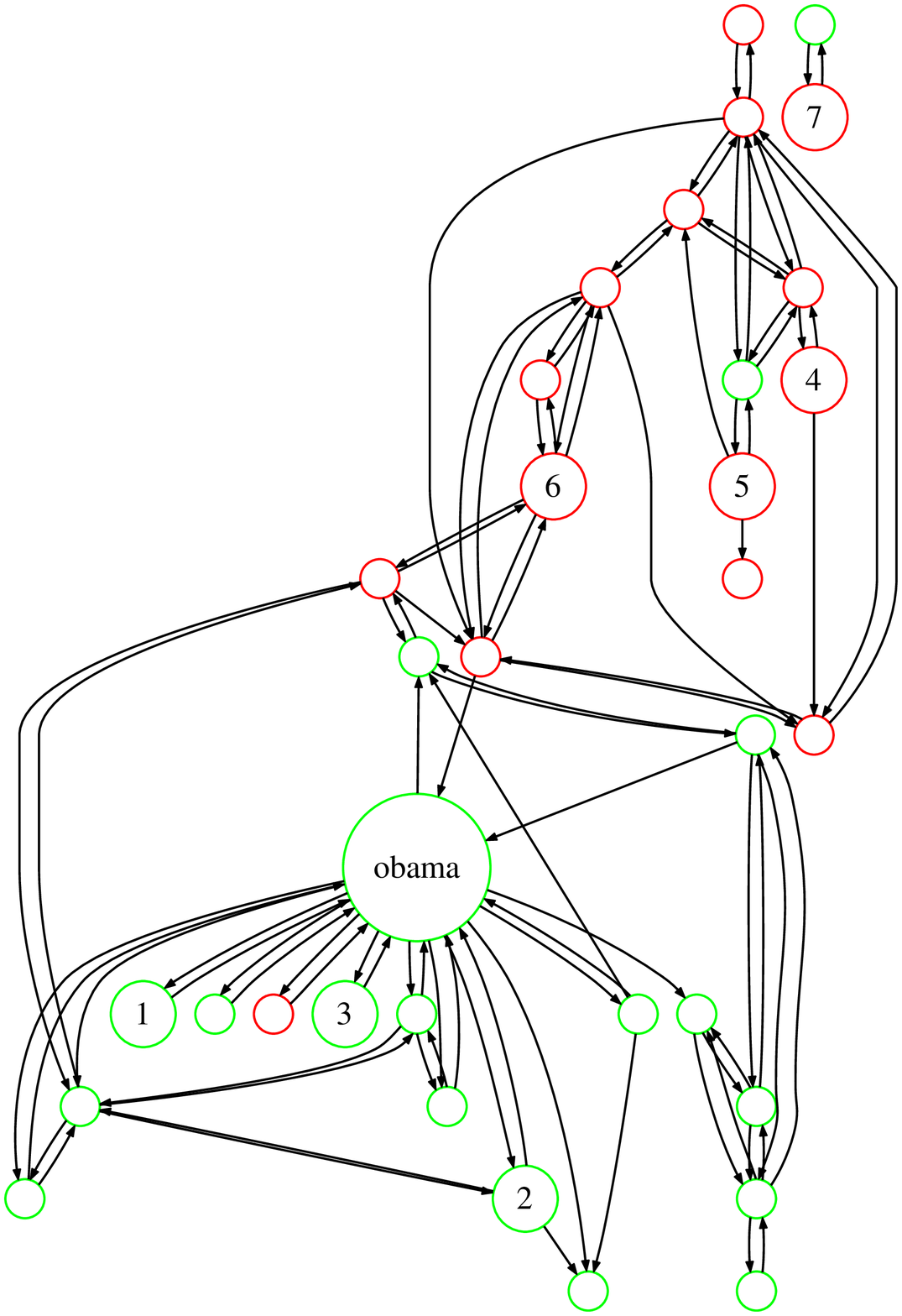,width=1.8in} &
\epsfig{figure=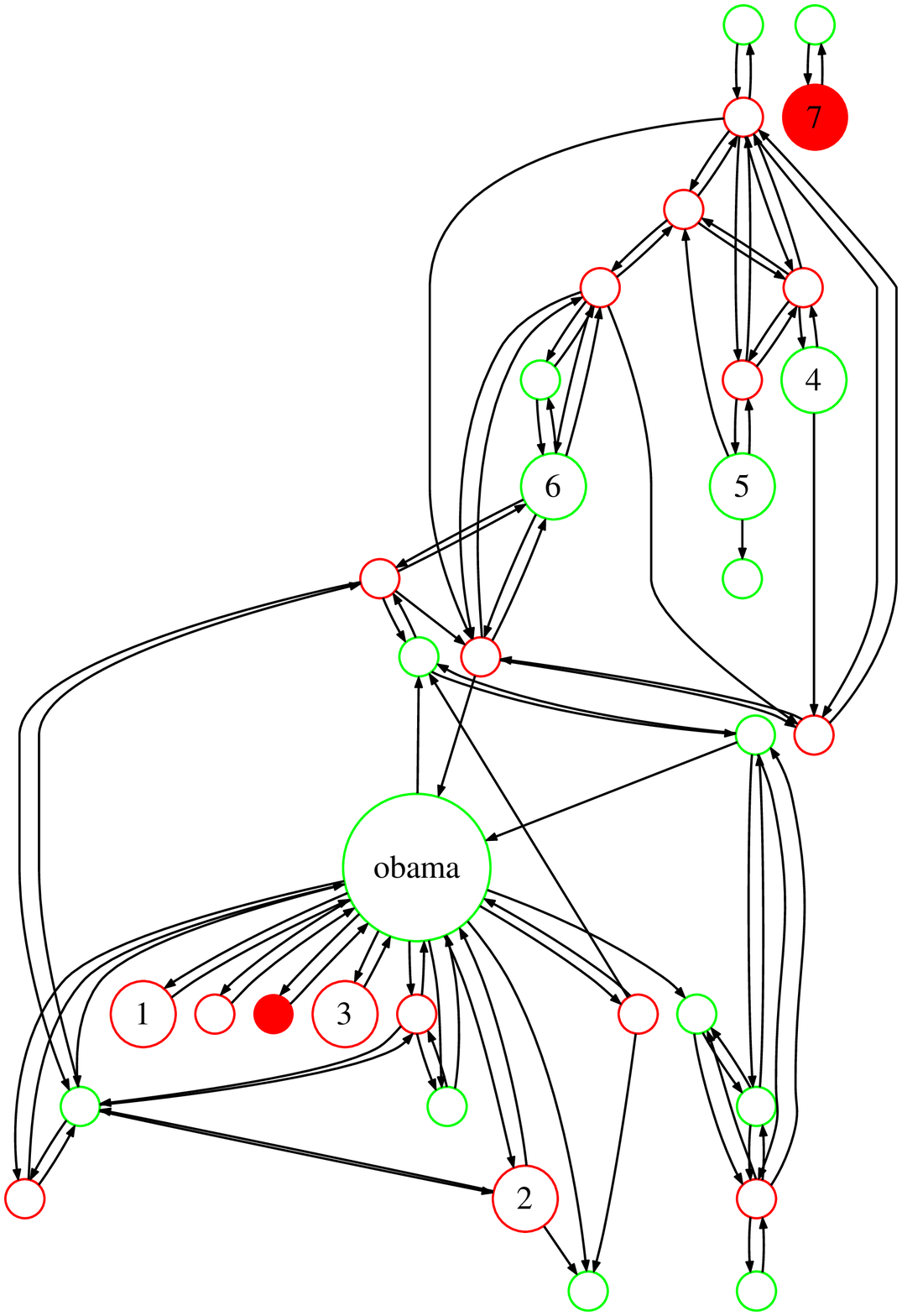,width=1.8in} &
\epsfig{figure=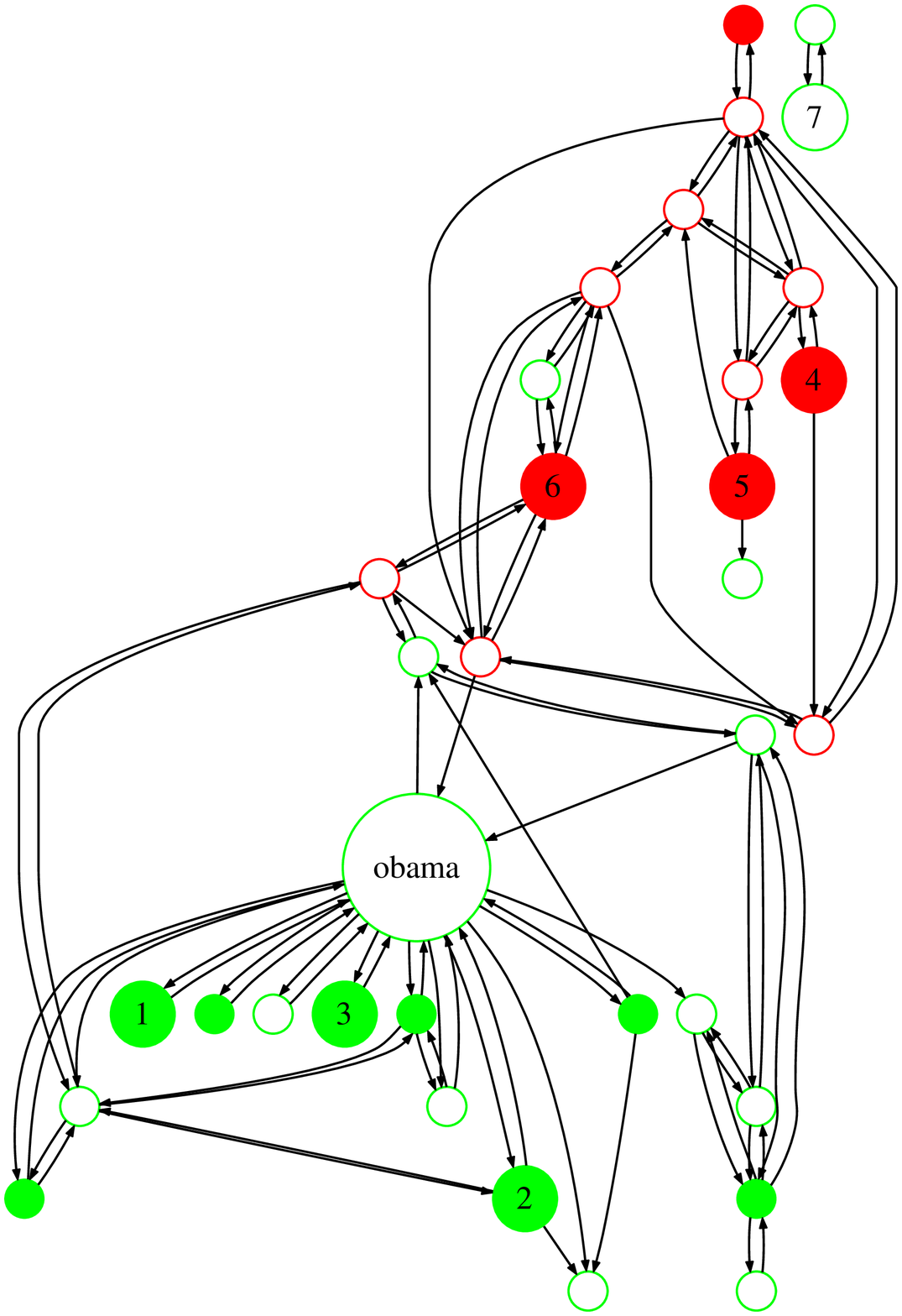,width=1.8in} \\
(a) Ground Truth & (b) Text-Only Approach & (c) Our algorithm
\end{tabular}
\end{center}
\caption{\label{fig:casestudy}Case study:
\small Portion of the \follow graph for the topic ``Obama'', where  derived labels on users are indicated by green (positive) and red (negative), respectively. 
Each node is a user, and the center one is ``BarackObama''.
The numbers in the nodes are indices into the table below.
 (a): Ground truth (human annotation).  (b) SVM Vote
(baseline).
(c) HGM-\srank{} in 
the
directed \follow graph.
Filled nodes indicate cases where the indicated algorithm was right
and the other algorithm was wrong; for instance, only our algorithm
was  correct on node 4.
}
\begin{center}
{\bf 
Sample tweets of users classified
correctly only when network information is incorporated}
\scriptsize
\begin{tabular}{|c|c|c|c|p{5.2in}|}
\hline
User ID & 
SVM Vote & HGM & True & Tweet \\ 
\hline
\hline

\multirow{1}{*}{1} & \multirow{1}{*}{NEG} & \multirow{1}{*}{POS} & \multirow{1}{*}{POS} & Obama is making the repubs look silly and petty. \#hrc 
\\
\hline 
\multirow{2}{*}{2} & \multirow{2}{*}{NEG} & \multirow{2}{*}{POS} & \multirow{2}{*}{POS} & Is happy Obama is President \\
\cline{5-5}
 & & & & Obama collectable http://tinyurl.com/c5u7jf \\ 
\hline 
\multirow{2}{*}{3} & \multirow{2}{*}{NEG} & \multirow{2}{*}{POS} & \multirow{2}{*}{POS} & I am praying that the government is able to get health care reformed this year! President Obama seems like the ONE to get it worked out!! \\
\cline{5-5}
 & & & & Watching House on TV. I will be turning to watch Rachel M. next. I am hoping Pres. Obama gets his budget passed. Especially Health Care! \\ 
\hline 
\multirow{2}{*}{4} & \multirow{2}{*}{POS} & \multirow{2}{*}{NEG} & \multirow{2}{*}{NEG} & RT @TeaPartyProtest Only thing we have 2 fear is Obama himself \& Pelosi \& Cong \& liberal news \& Dems \&... http://ow.ly/15M9Xv \\
\cline{5-5}
 & & & & RT @GlennBeckClips: Barack Obama can no more disown ACORN than he could disown his own grandmother. \#TCOT \\ 
\hline 
\multirow{2}{*}{5} & \multirow{2}{*}{POS} & \multirow{2}{*}{NEG} & \multirow{2}{*}{NEG} & RT @JosephAGallant Twitlonger: Suppose I wanted to Immigrant to Mexico? A Letter to President Obama.. http://tl.gd/1kr5rh \\
\cline{5-5}
& & & & George Bush was and acted like a war time President. Obama is on a four year power grab and photo op. \#tcot \\ 
\hline 
\multirow{2}{*}{6} & \multirow{2}{*}{POS} & \multirow{2}{*}{NEG} & \multirow{2}{*}{NEG} & ObamaCare forces Americans to buy or face a fine! It is UNCONSTITUITIONAL to force us to buy obamacare. Marxist Govt. taking our Freedoms! \\
\cline{5-5}
& & & & Look up Chicago Climate Exchange,an organization formed years ago by Obama \& his Marxist-Commie Cronies to form a profit off  cap \& trade. \\ 
\hline 
\end{tabular}
\normalsize

\end{center}
\end{figure*}

In this section, we first describe our experimental procedure.  
We then present a case study 
that validates our intuitions as to
how the network-structure information helps user-level sentiment classification. Finally,  
we analyze the performance results,  
and examine the effects of  graph density, edge ``quality'', SVM training data, and amount of unlabeled data.

\subsection{Experimental Procedure} \label{sec:expproc}
We ran each experiment 10 times.
In each run, 
we partitioned
the data for which we had ground-truth labels into a training set,
consisting of 50 positive plus 50 negative randomly-chosen users whose
labels are revealed to the algorithms under consideration, and an
evaluation set consisting of the remaining \labeled users.
\footnote{Note that the ratios of $|training~set|/|evaluation set|$
  are different in different topics.}

One issue we have not yet addressed is our complete lack of
annotations on the tweets; we need tweet labels as part of our model.
We construct training data 
where the ``positive'' tweets were 
all (on-topic) tweets from users \labeled positive,
and the ``negative'' tweets
all (on-topic) tweets from users
\labeled negative.
(We discuss some alternative approaches later.)
Different classifiers are trained for different topics.

We 
compare three user-classification methods,
two of which
were introduced in 
\S\ref{sec:approach}
and the other of which is  our baseline:
\begin{itemize}
\setlength{\itemsep}{-0.2\baselineskip}
\item %
\textit{Majority-vote Baseline (SVM Vote)}: The user's sentiment label
  is simply the majority label among their (on-topic) tweets,
according to the SVM.
\footnote{We also tried the baseline 
of combining all the 
(on-topic)
tweets of a
 user into a single document; 
the 
results were much worse.
}

\item \textit{Heterogeneous Graph Model with \stalong (HGM-\sta)}: We
derive parameter values
according to the statistics in
the
 labeled data, and then 
apply loopy belief propagation to
 infer 
user sentiment labels.
\item \textit{Heterogeneous Graph Model with \sranklong (HGM-\srank)}:
  We 
perform
semi-supervised learning on the heterogeneous graph and then apply loopy belief propagation to get user-level sentiment labels. 
\end{itemize}

We measure performance via both accuracy and Macro F1 on the 
evaluation set.

\newcommand{\topicheader}[1]{{\bf #1}}

\subsection{Case Study}

We first engage in a 
case  study to show how
our graph information can improve
 sentiment analysis. 
Figure \ref{fig:casestudy} shows an example generated from our experiments.
In the 
depicted portion (a)  of the 
ground-truth user graph for the topic ``Obama'', we see that positive (green) and negative (red) users are relatively clustered.
Deriving user labels from an SVM run on text alone yields graph (b), in which we see much less clustering and a number of mistakes compared to ground truth: 
this is probably because 
tweet text is short and relatively difficult to interpret, according to our initial inspections of the data.  
In contrast, graph (c)
shows that our text- and network-aware algorithm produces labels that are
more coherently clustered and 
correct more often 
than (b).

We investigate more by looking at 
a specific example. 
The table in the lower part of Figure \ref{fig:casestudy}
shows 
a selection of  
tweets for users
that only our algorithm classified correctly.
We see that the text of
these 
tweets 
is often seemingly hard (for an SVM) to classify.  For example,
user 1's ``Obama is making the repubs look silly and petty. \#hrc''
has negative words in it, although it is positive towards Obama. 
In these cases, 
the network structure 
may connect
initially-misclassified
users to users
with the same sentiment, and our network-aware algorithm is able to
use such relationship information to overcome the difficulties of
relying on text interpretation alone.

It should be pointed out that there are cases where text alone is a better source of information.
Consider user 7 in Figure \ref{fig:casestudy}, who resides in a two-node
connected component and was correctly classified by SVM Vote but not
HGM-\srank{}.  User 7 is particularly prolific, so there is a great
deal of data for the text-based SVM to work with; but the
network-based method forced user 7 to share its neighbor's label
despite this preponderance of textual evidence.

\subsection{Performance Analysis}
\begin{figure}[htb!]
\begin{center}
\scriptsize
\begin{tabular}{cc}
Accuracy & MacroF1\\
\epsfig{figure=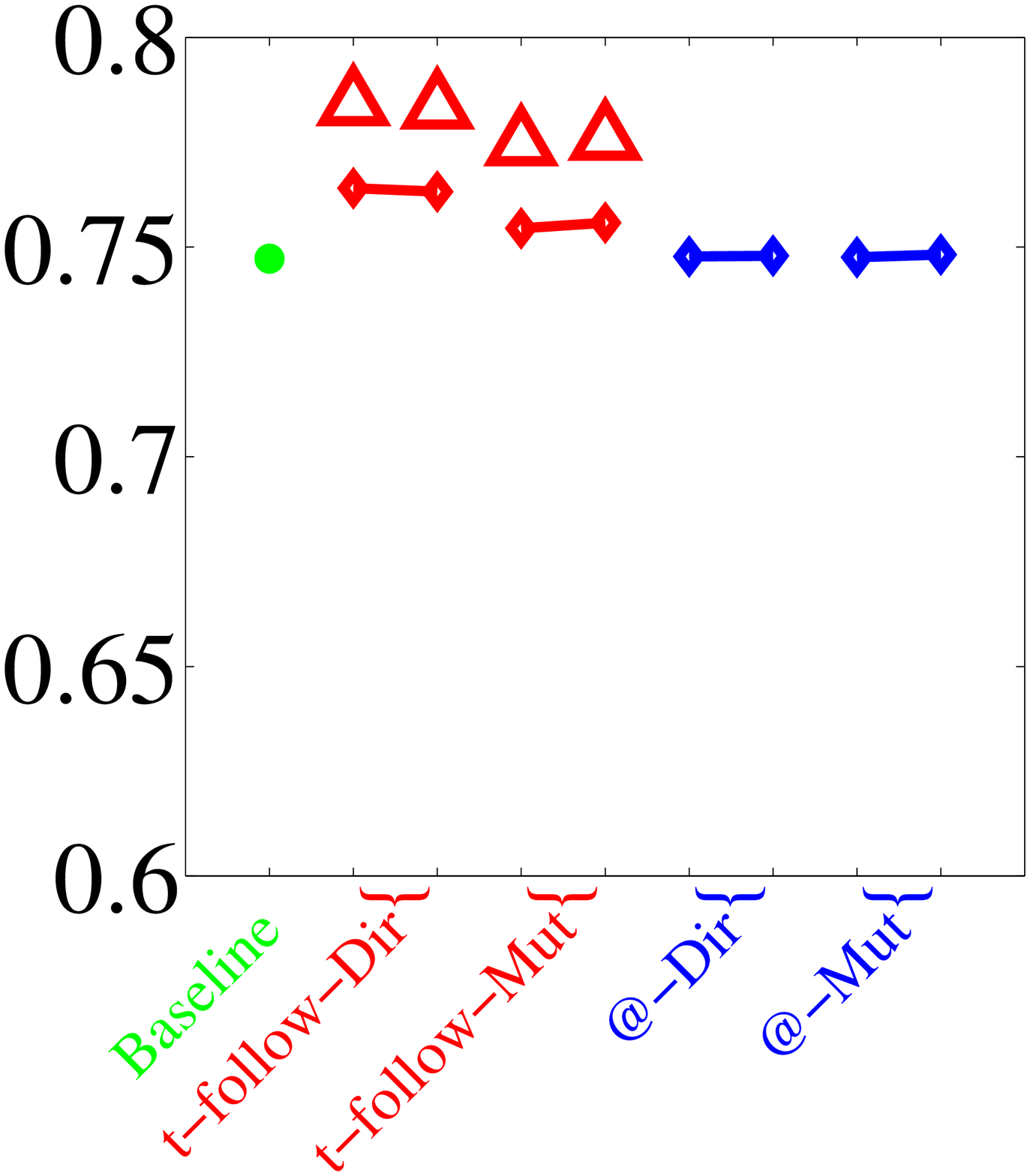, width=1.6in} & 
\epsfig{figure=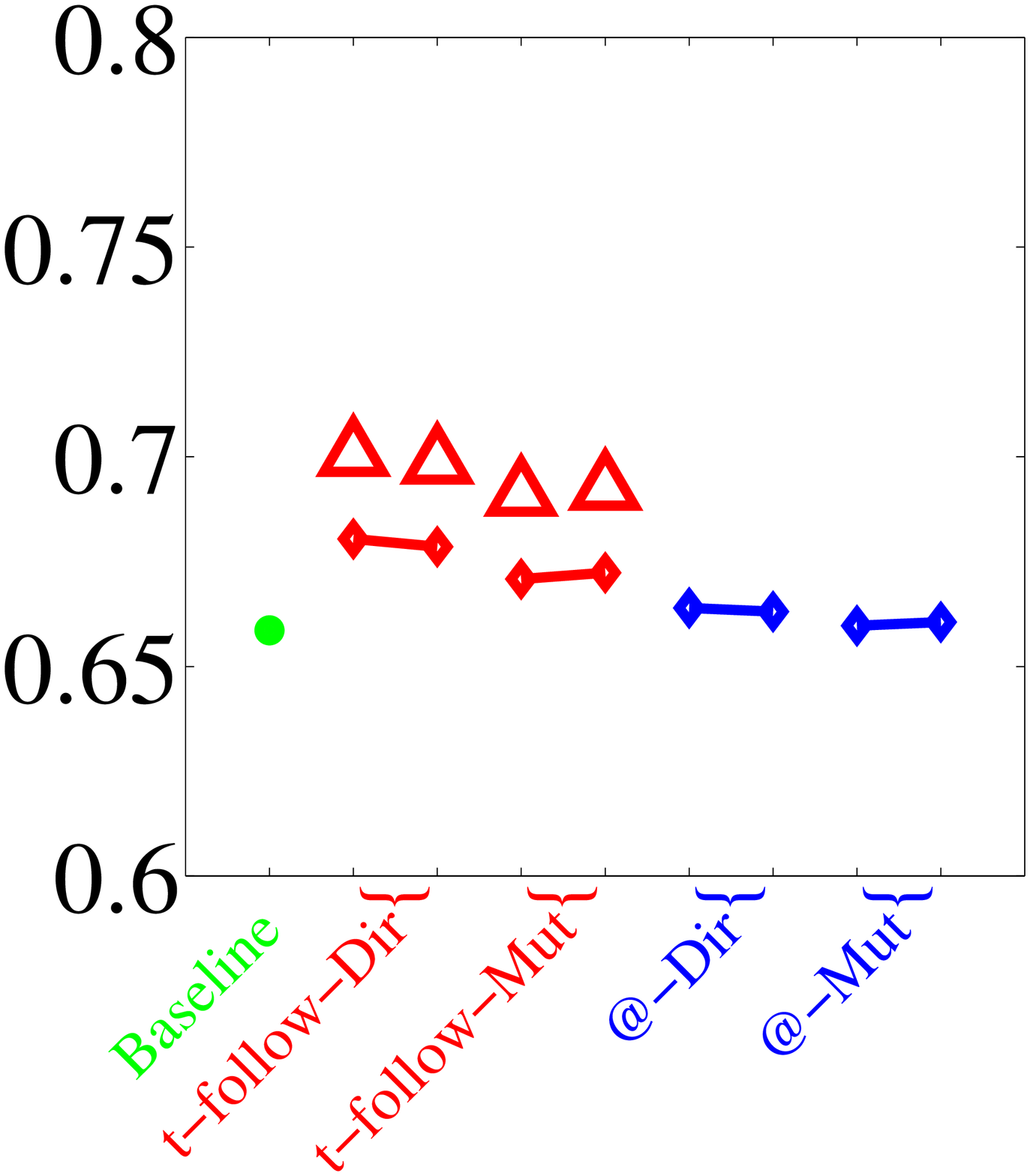, width=1.6in}
\end{tabular}
\end{center}
\caption{\label{fig:exampleperf}Average Performance Analysis. \small
Red indicates \follow graphs, blue indicates @ graphs.
 For each connected pair, the left one is from \sta, while the right
  one is from \srank. 
A
$\triangle$ 
marks those approaches that are 
significantly better than
the baseline for more
 than 3 topics.
}
\normalsize
\end{figure}

\begin{figure*}[htb!]
\begin{center}
\scriptsize
\begin{tabular}{ccccc}
\multicolumn{5}{c}{\topicheader{Accuracy}} \\
Obama & Sarah Palin & Glenn Beck & Lakers & Fox News\\
\epsfig{figure=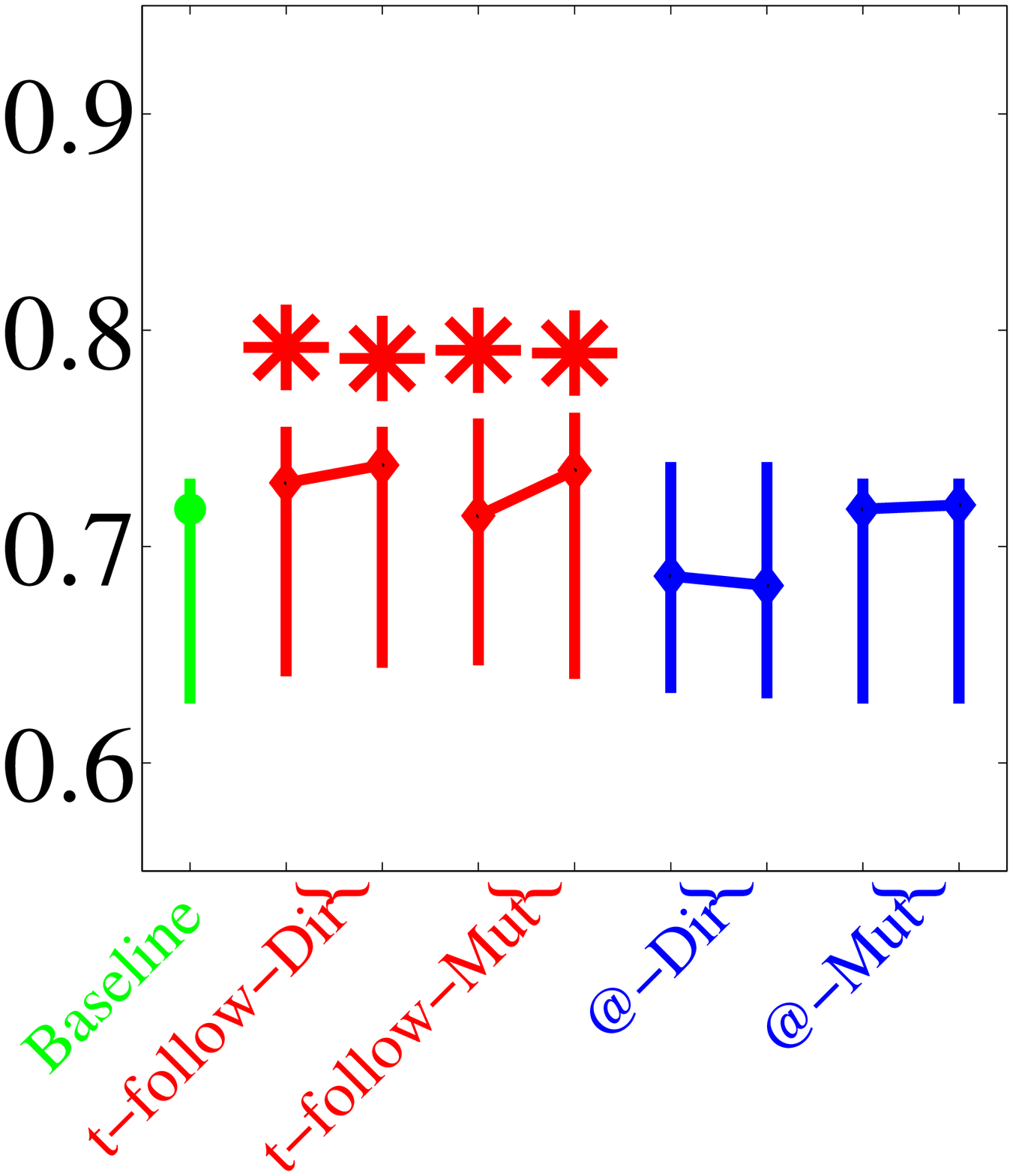, width=1.2in} &
\epsfig{figure=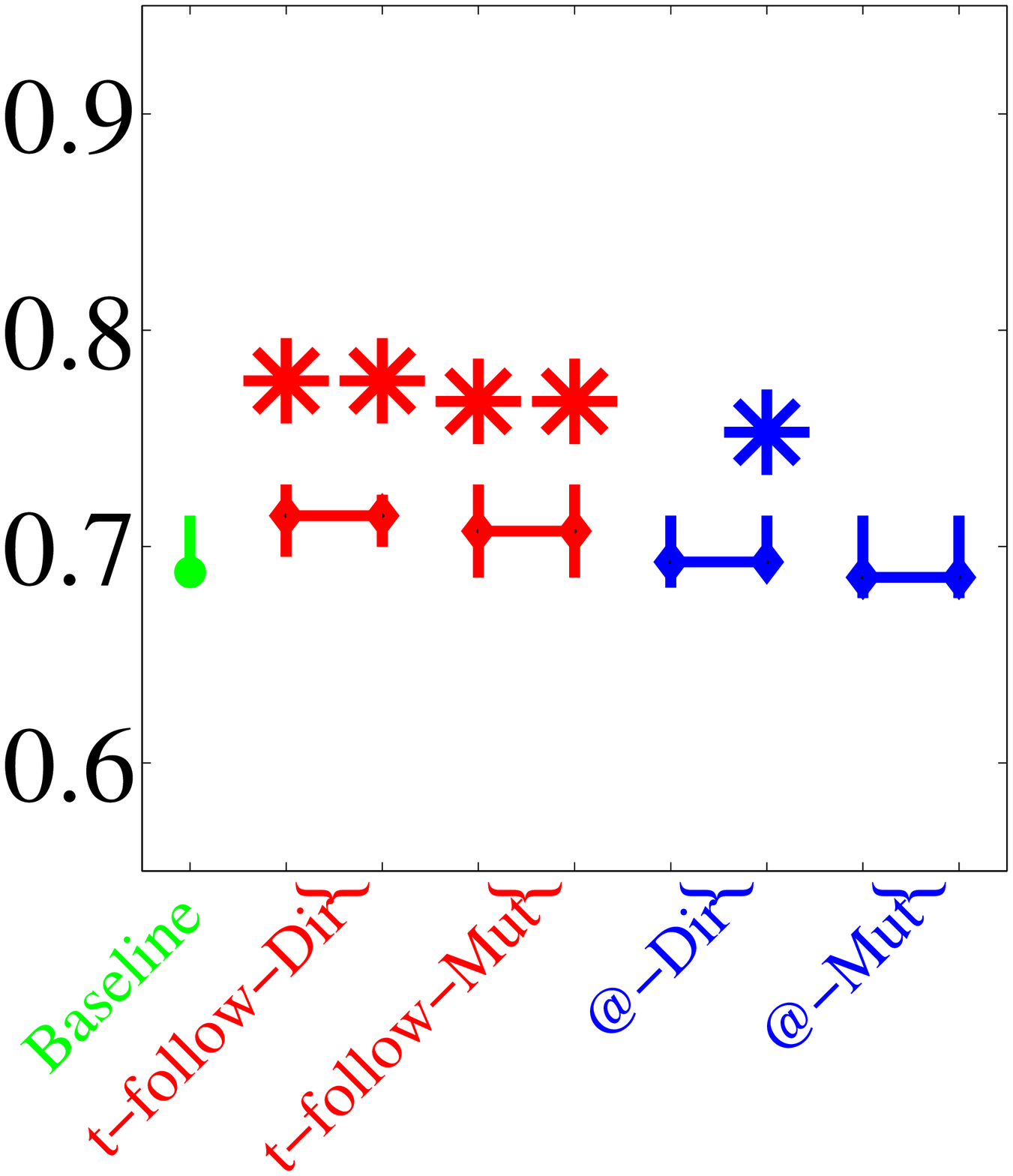, width=1.2in} &
\epsfig{figure=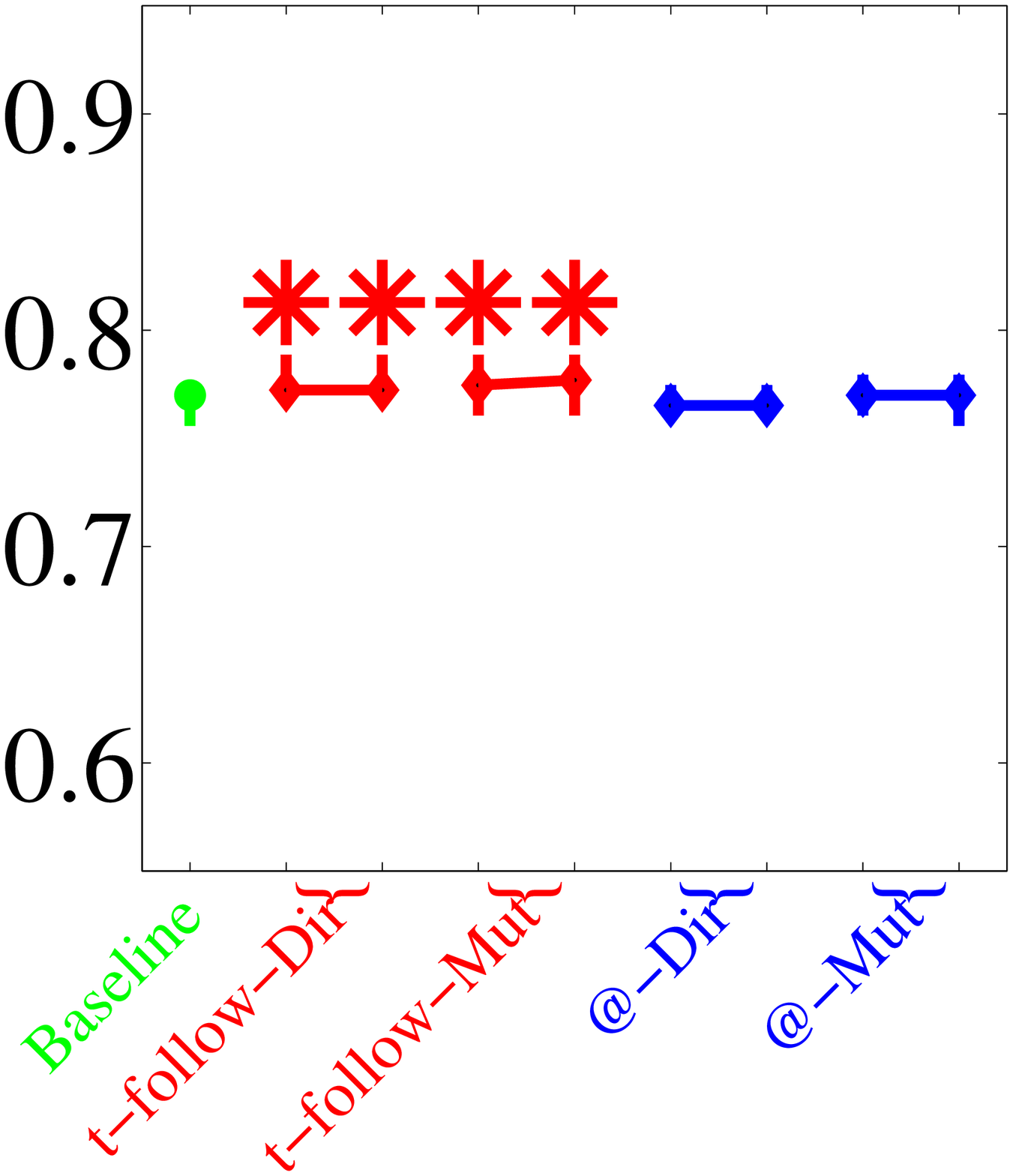, width=1.2in} &
\epsfig{figure=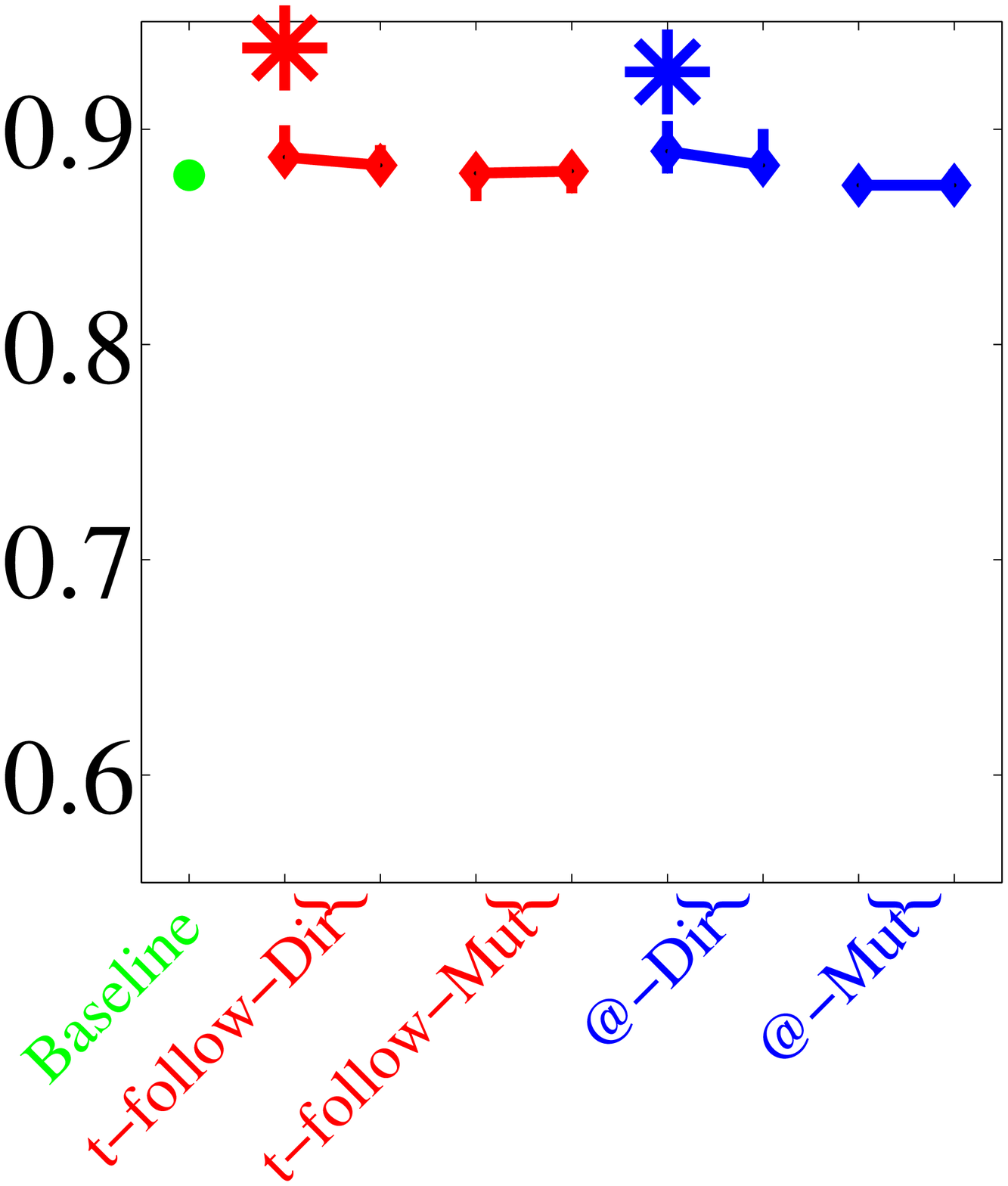, width=1.2in} &
\epsfig{figure=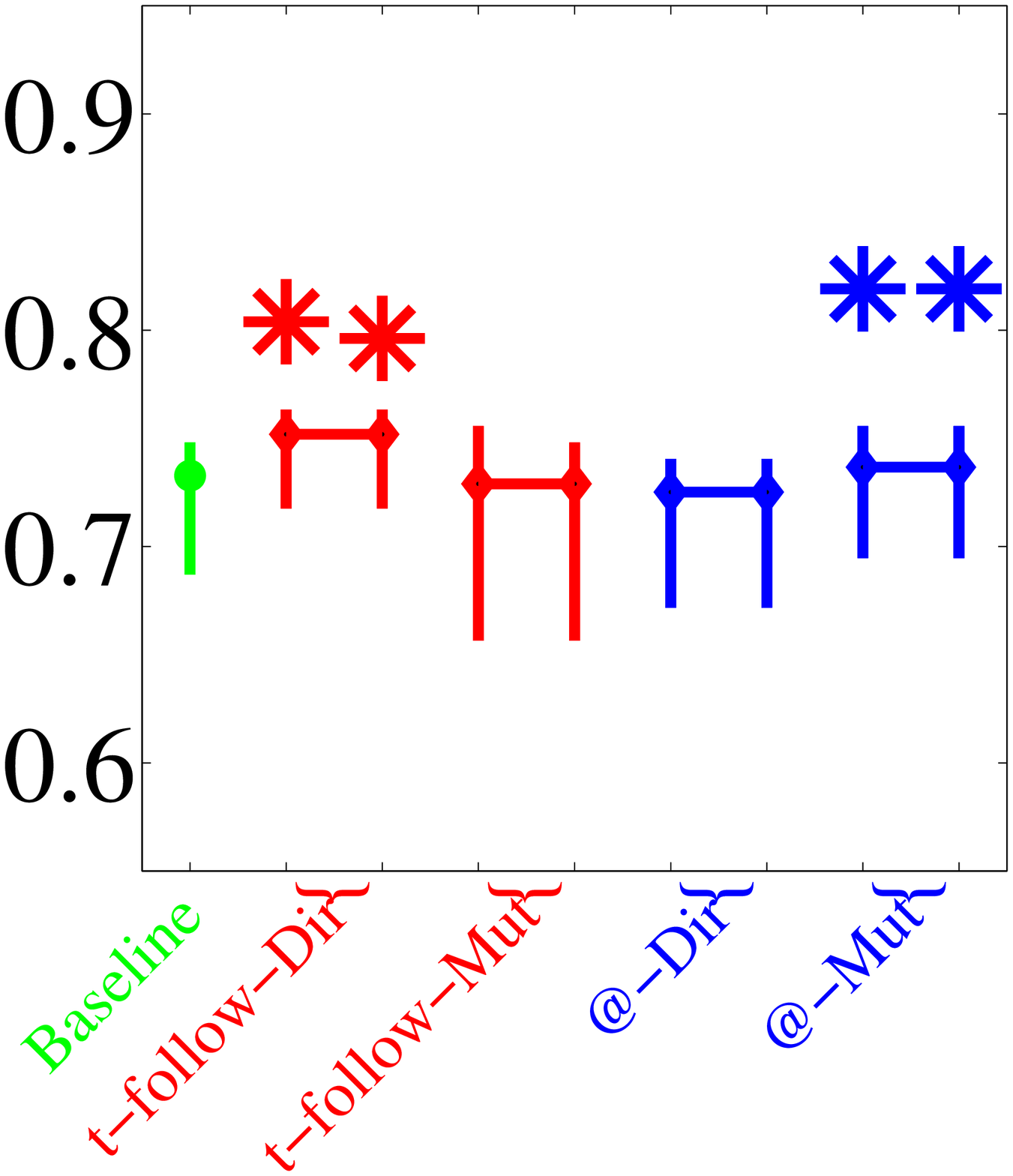, width=1.2in}\\
\multicolumn{5}{c}{\topicheader{MacroF1}} \\
Obama & Sarah Palin & Glenn Beck & Lakers & Fox News\\
\epsfig{figure=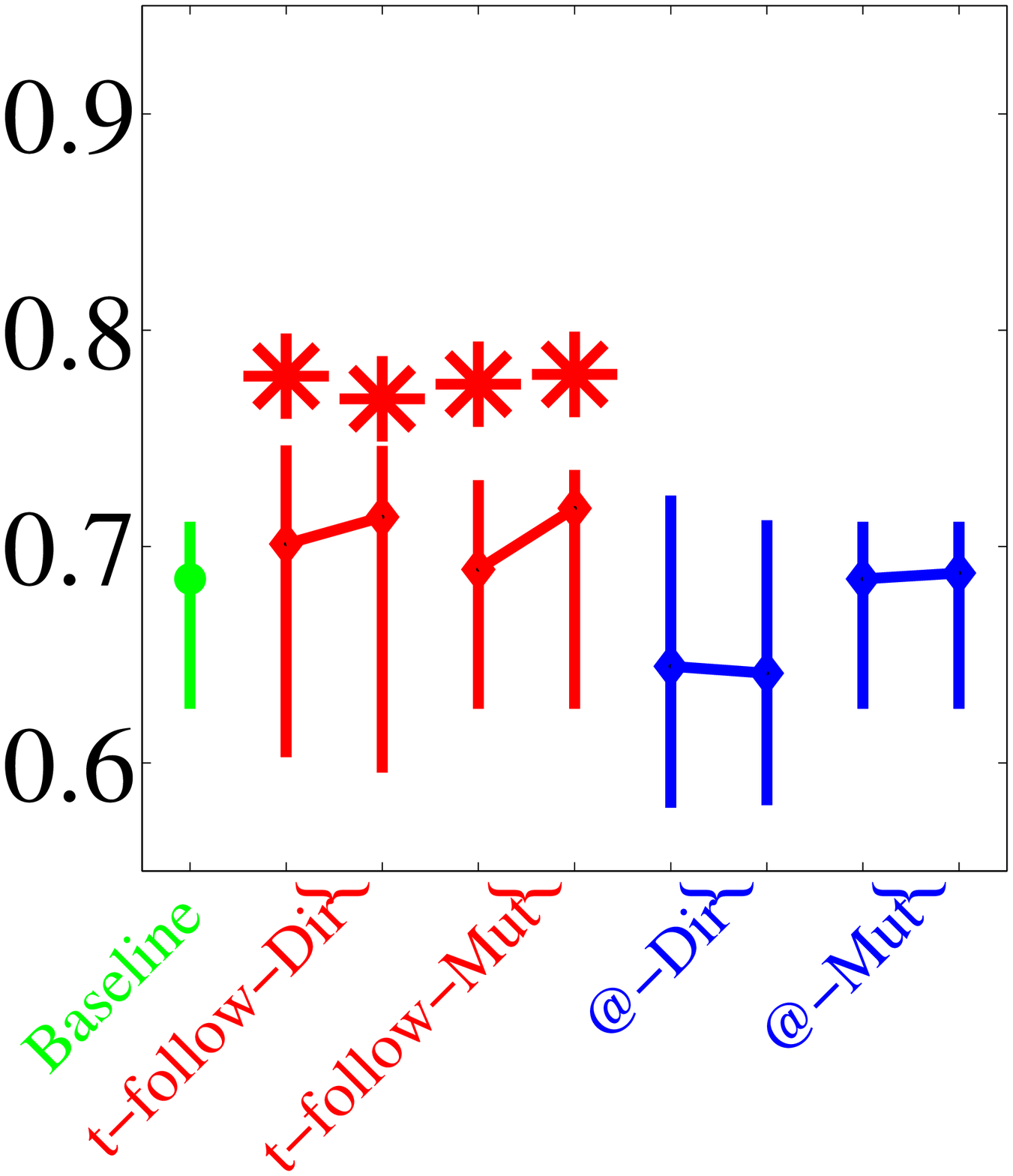, width=1.2in} &
\epsfig{figure=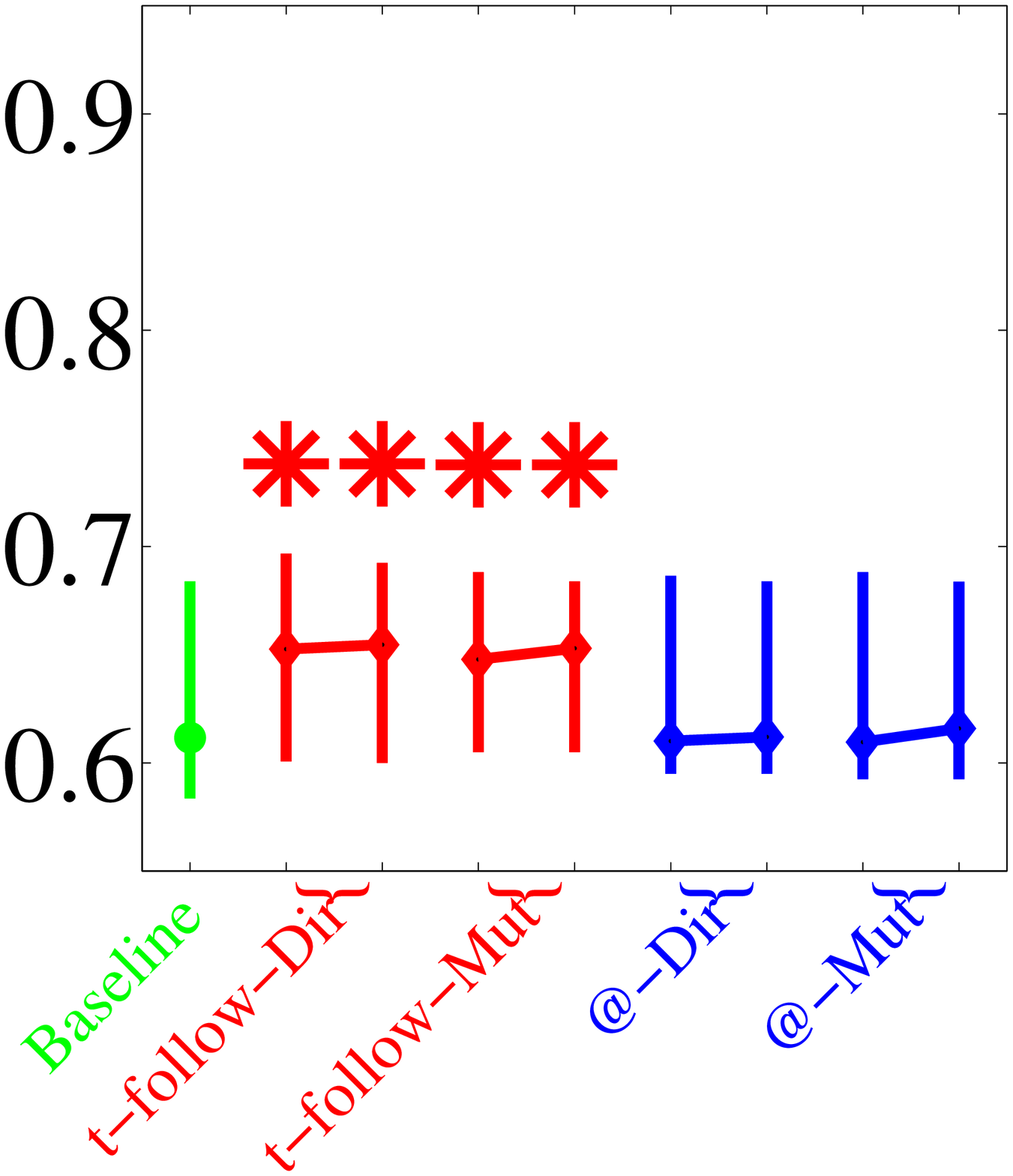, width=1.2in} &
\epsfig{figure=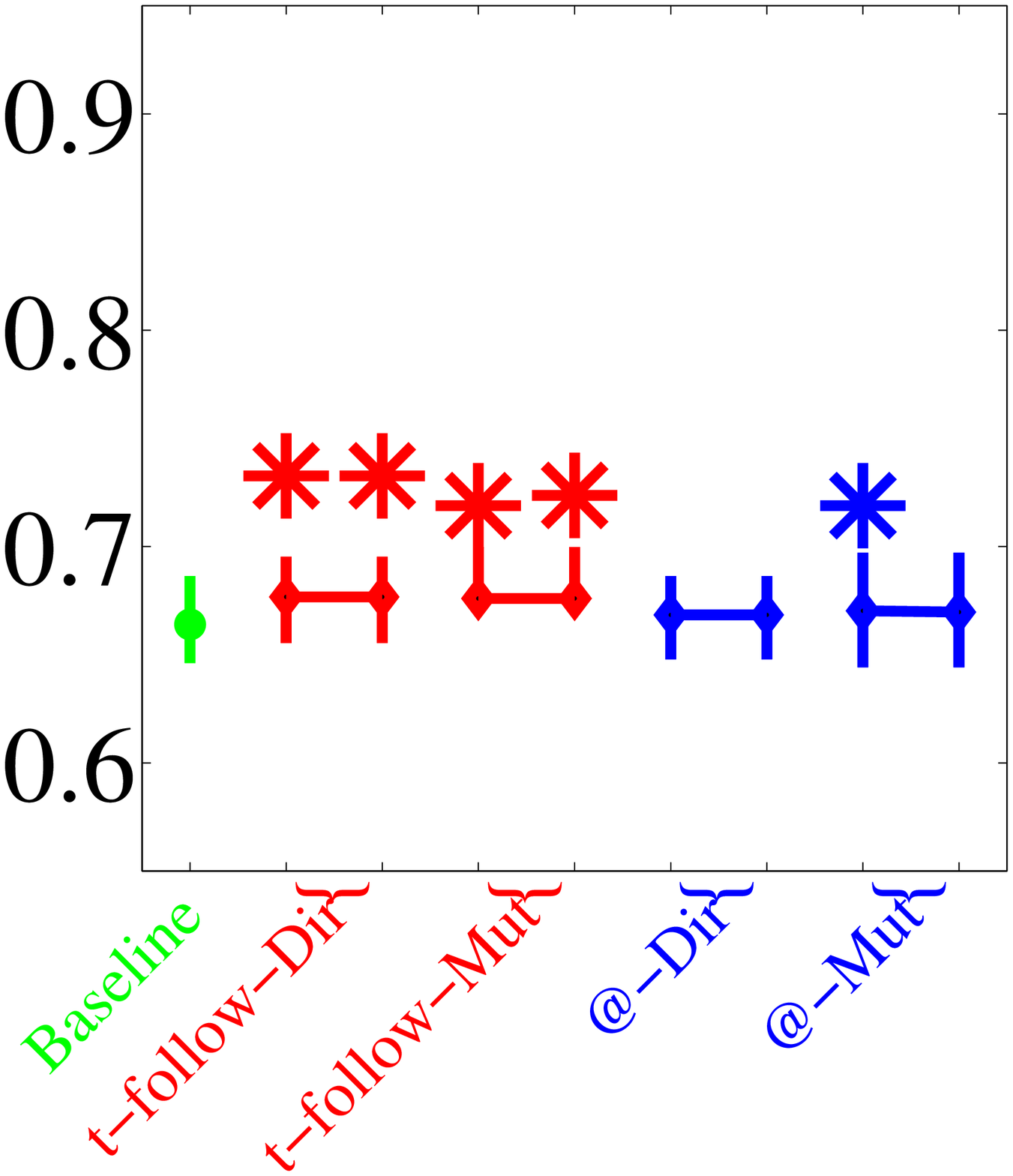, width=1.2in} &
\epsfig{figure=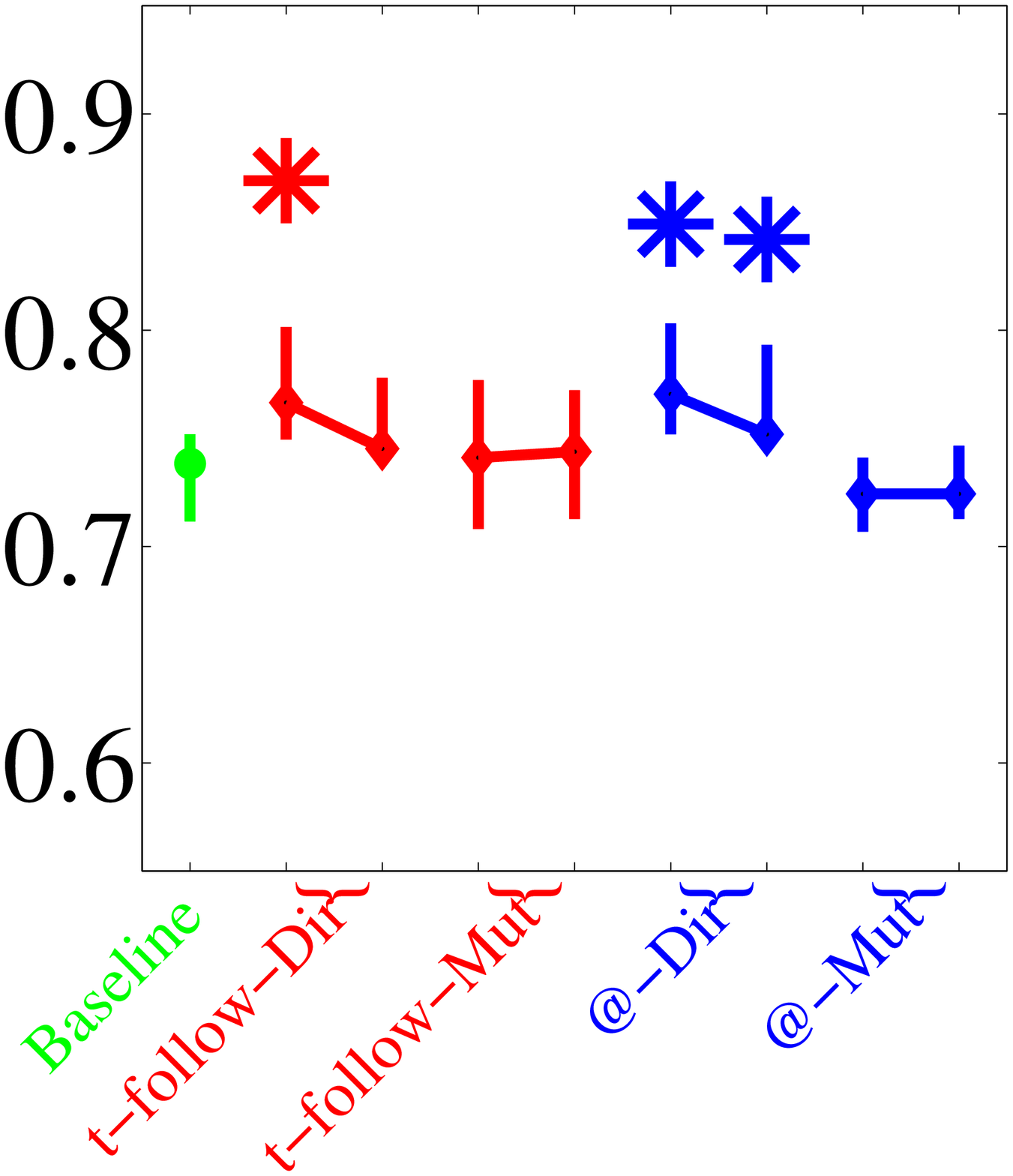, width=1.2in} &
\epsfig{figure=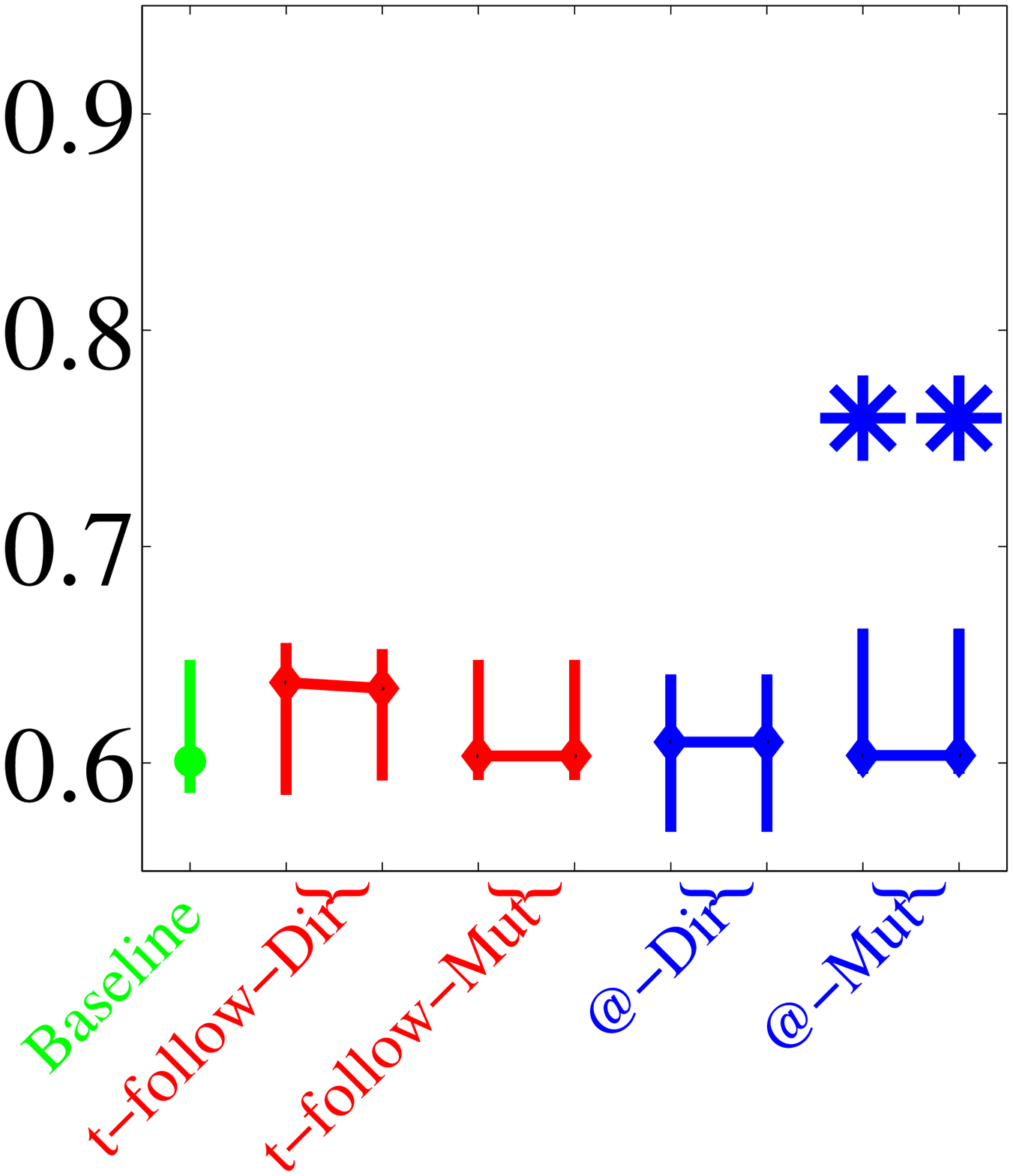, width=1.2in}\\
\end{tabular}
\end{center}
\caption{\label{fig:expperf}Performance Analysis in Different Topics. \small 
The 
x-axes are the same as in Figure \ref{fig:exampleperf}.
Bars summarize performance results for our ``10-run'' experiments:
the bottom and top of a bar indicate the 25th and 75th percentiles,
respectively. 
Dots indicate median results;  in pairs connected by lines, the left is  ``\sta'', while the right is ``\srank''.
Green: SVM vote, our baseline.  Red: network-based approaches applied to the \follow  graphs. Blue: results for the @ 
graphs.  
Stars ($\ast$) indicate
performance that is significantly better than the baseline, according
to the paired t-test.}
\normalsize
\end{figure*}

We now present the performance
results
 for the different methods we
considered. Figure \ref{fig:exampleperf} shows the average performance
of 
the
different methods across topics. The green dot represents the
performance of
the
 baseline, the red ones are results 
for \follow graphs, 
and the blue ones are results
for @ graphs.
The presence of a 
$\triangle$
indicates
that the corresponding approach is significantly better than
the 
baseline 
for more than 3 topics.

First, our approaches all show better performance than the baseline
both in Accuracy and MacroF1, though 
the improvement is rather small
in @ graphs. This validates the effectiveness of incorporating network information. 

Second, \follow graphs 
(red)
show better performance  than @ graphs
(blue). 
It seems that \follow
 relations between people are 
more reliable 
indicators of sentiment similarity, which is consistent with our 
analysis of Figure \ref{fig:linkStatistics}.

Third, directed graphs work better than \undirected graphs. 
This could either be because approval/attention links are more related
to shared sentiment than any effects due to homophily, or because the
directed graphs are denser than the \undirected ones, as can be seen
from 
Table \ref{tb:sparsitystatistics}.

Fourth, 
\sta and \srank performed quite similarly.  (However, we show below that \srank can provide more robustness when more unlabeled users are added.)

\begin{figure*}[htb!]
\begin{center}
\begin{tabular}{ccccc}
Obama & Sarah Palin & Glenn Beck & Lakers & Fox News \\
\epsfig{figure=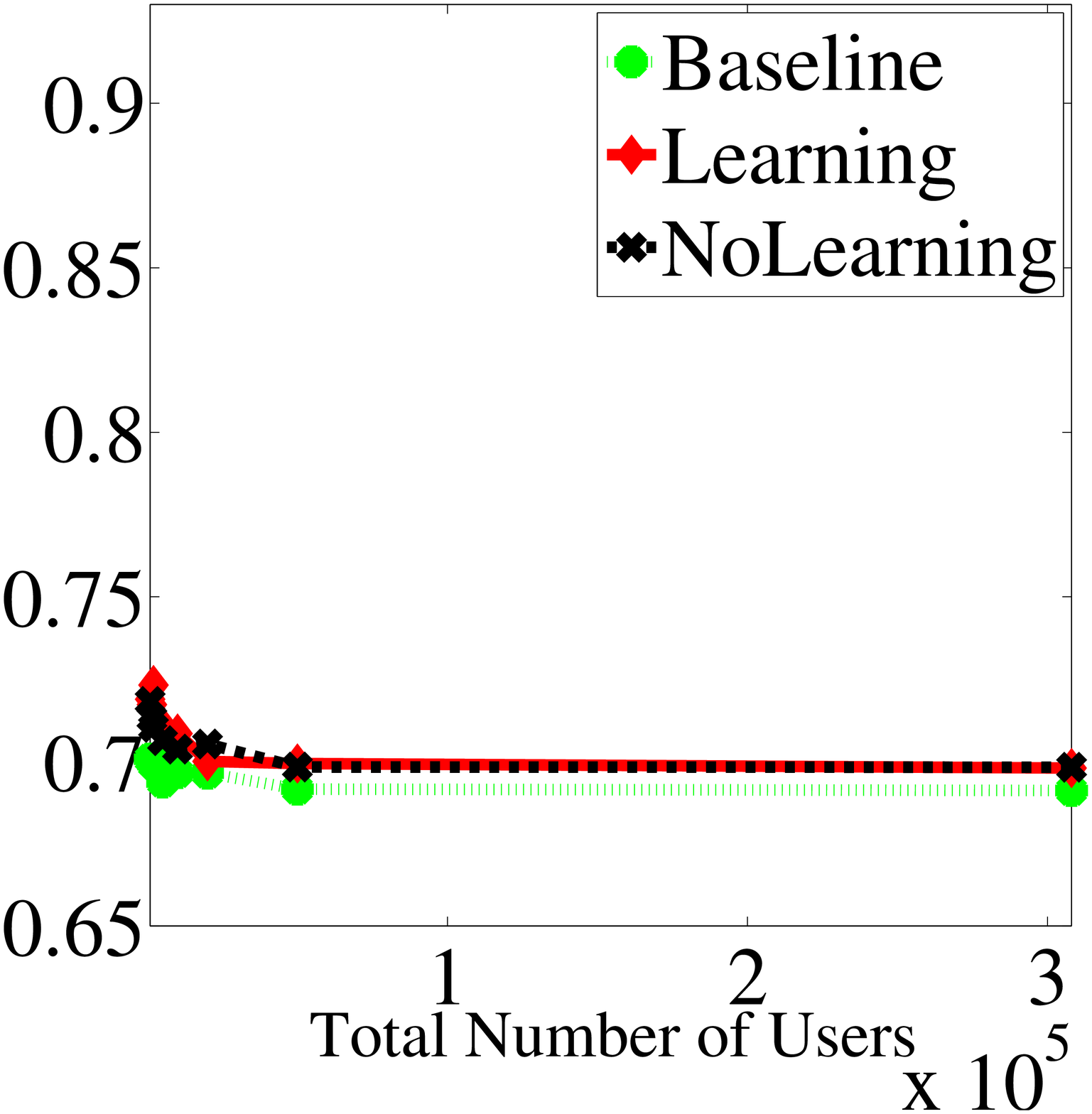, width=1.2in} &
\epsfig{figure=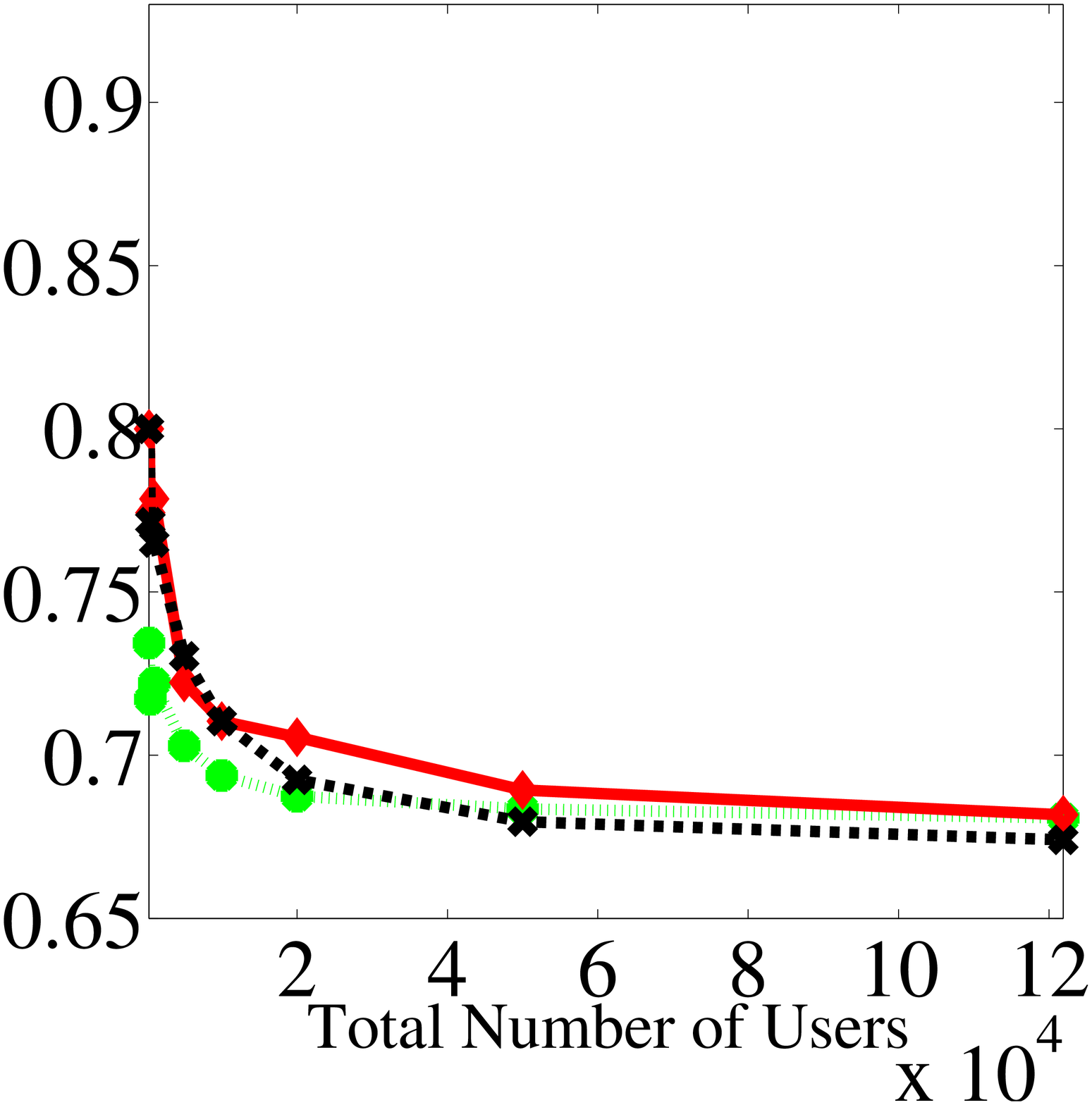, width=1.2in} &
\epsfig{figure=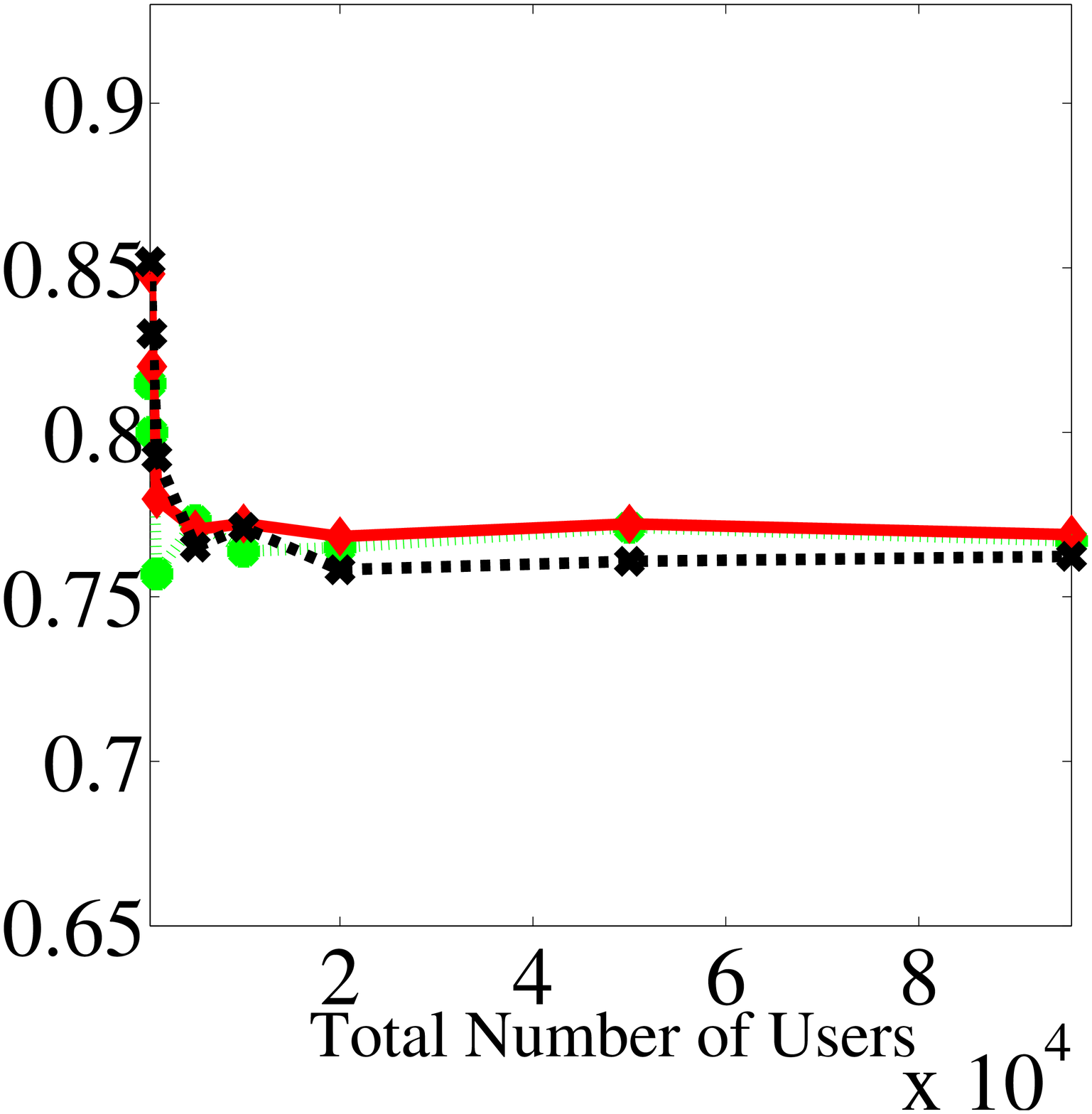, width=1.2in} &
\epsfig{figure=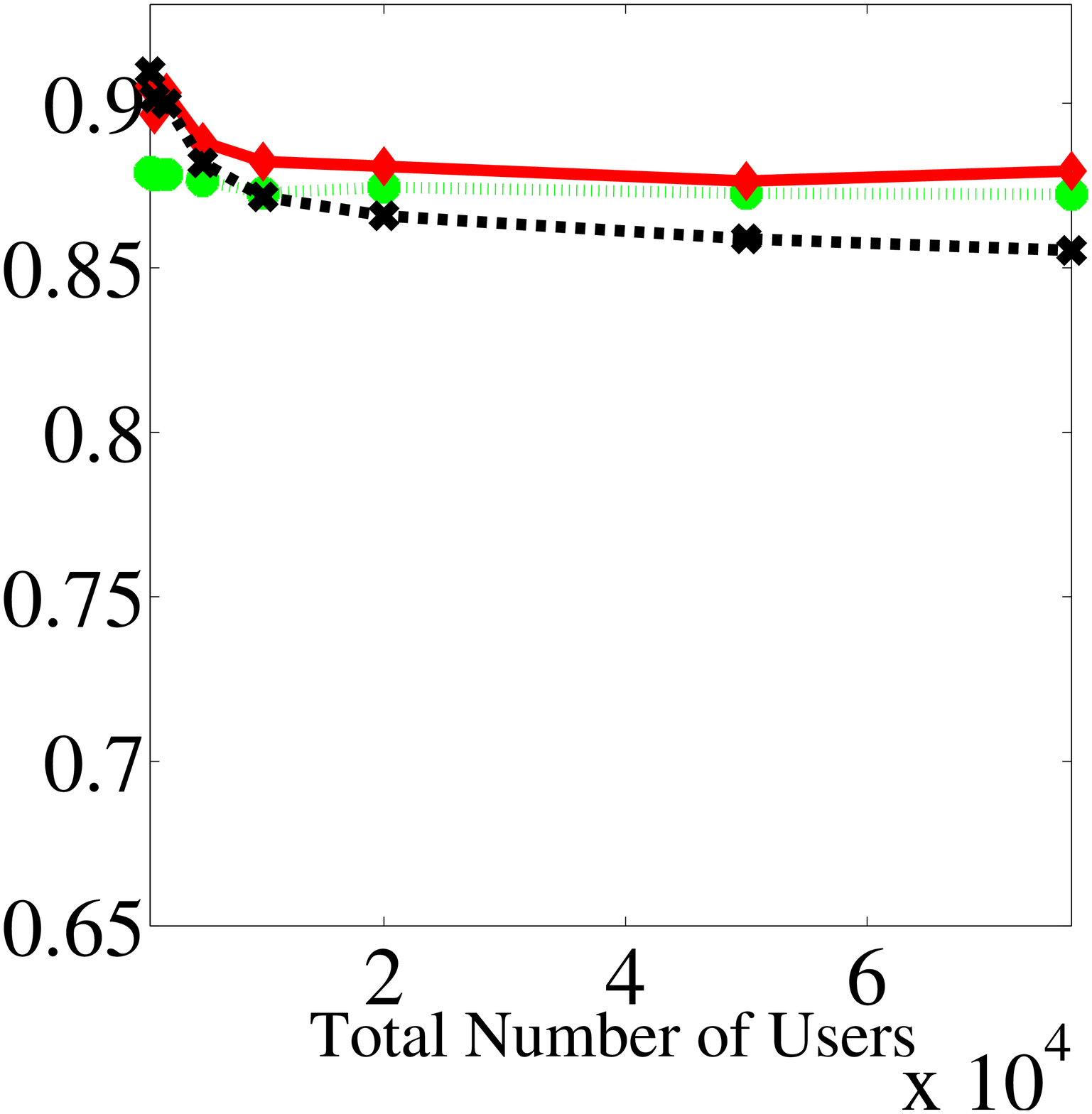, width=1.2in} &
\epsfig{figure=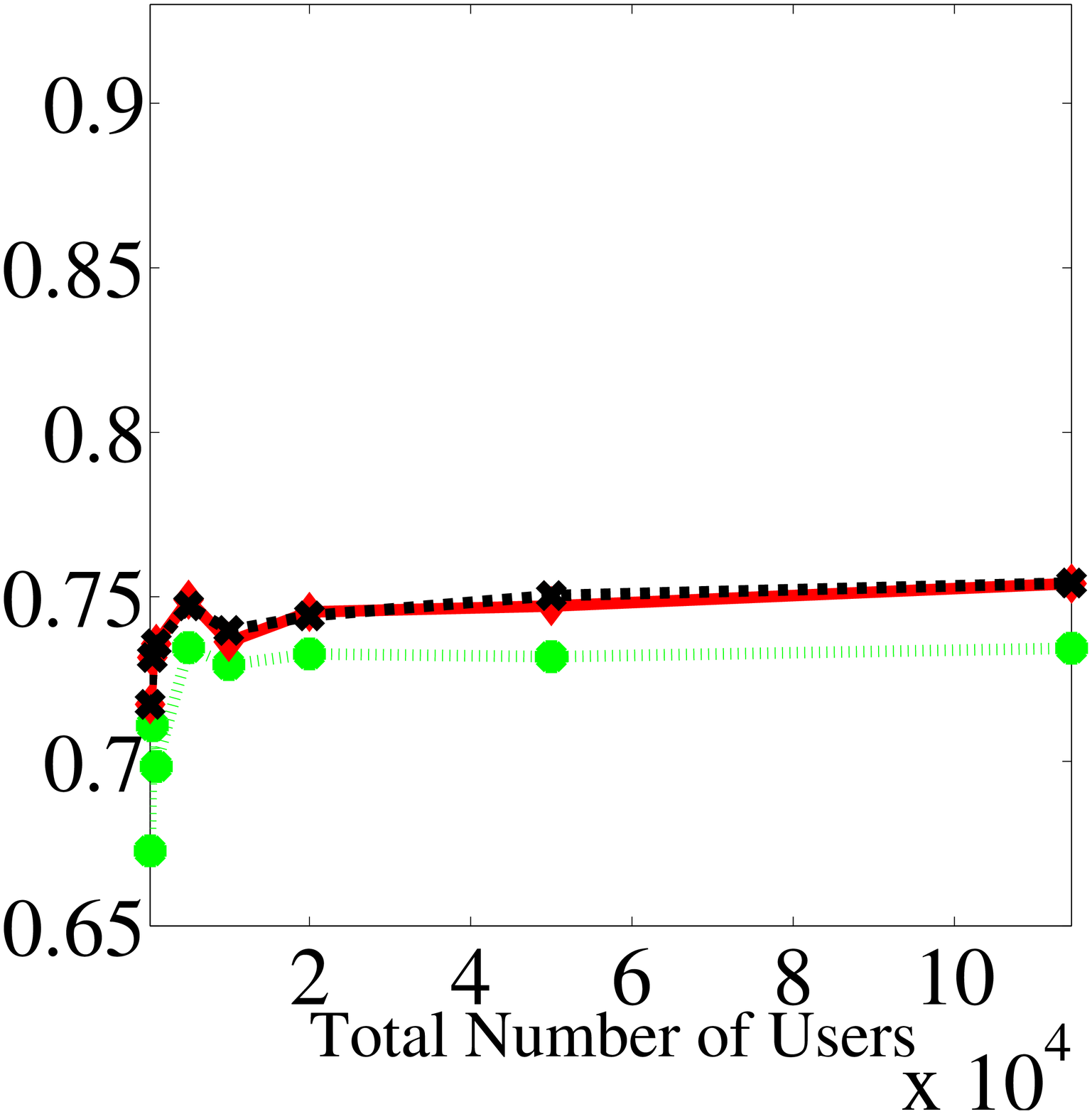, width=1.2in}
\end{tabular}
\end{center}
\caption{\label{fig:unlabeledperf} Accuracy in the Largest Connected
  Component. \small We show the average accuracy 
 in the
  largest connected component of
the
 directed \follow graph
 as the amount of \unlabeled data increases.}
\end{figure*}
\vpara{Per-topic performance: density vs. quality analysis}
We now look at the topics individually to gain a better understanding
of what factors affect performance.  Figure \ref{fig:expperf} gives
the per-topic breakdown.  
Again,
we use green, red, and blue to indicate, respectively,  the SVM-vote baseline, our graph-based methods using \follow graphs, and our graph-based methods using @ graphs. 
The $\ast$'s denote where our approach is significantly better than the baseline
(paired t-test, .05 level). 
Overall, we see that for the topics  ``Obama'', ``Sarah Palin'' and
``Glenn Beck'', the \follow graph is much more effective than the @
graph in terms of providing statistically significant improvements
over the baseline; but for the topics ``Lakers'' and ``Fox'', the @ graph provides more instances of statistically significant improvements, and overall there are fewer statistically significant improvements over SVM vote.  What accounts for these differences?

\begin{table}[htb!]
\centering \caption{\label{tb:sparsitystatistics} 
Average degree 
statistics.
Directed degree refers to
out-degree.}
\scriptsize
\begin{tabular}{|c||r|r|r|r|r|}
\hline 
\emph{Topic} & \# users & \multicolumn{2}{c|}{\follow graph} &  \multicolumn{2}{c|}{@ graph} \\ \cline{3-6}
                      &                           & \multicolumn{1}{c|}{\anylong} & \undirected   & \multicolumn{1}{c|}{\anylong} & \undirected \\ 
\hline \hline Obama & 889 & 8.8 & 6.6 & 2.7 & 0.7\\
\hline Sarah Palin & 310 & 3.2 & 1.7 & 1.4 & 0.4\\
\hline Glenn Beck &  313 & 1.6 & 1.0 & 0.5 & 0.1\\
\hline Lakers & 640 & 3.6 & 1.1 & 1.8 & 0.4\\
\hline Fox News & 231 & 0.6 & 0.3 & 0.2 & 0.04\\
\hline
\end{tabular}
\normalsize
\end{table}

\begin{table*}[htb!]
\centering \caption{\label{tb:fullgraphstat} 
Statistics on the 
expanded graphs.  
Boldface indicates the setting used in Figure \ref{fig:unlabeledperf}.
}
\scriptsize
\begin{tabular}{|c|r||r|r|r|r||r|r|r|r||r|}
\hline 
\emph{Topic} & \# users & \multicolumn{2}{c|}{\# \follow edges} &  \multicolumn{2}{c||}{\#@ edges}  & \multicolumn{2}{c|}{average \follow degree} & \multicolumn{2}{c||}{average @ degree} & total \# of on-topic tweets \\ \hline
                      &                           &
                     \multicolumn{1}{c|}{{\bf \anylong}} &
                      \undirected   &
                     \multicolumn{1}{c|}{\anylong} & \undirected
                     &\multicolumn{1}{c|}{\bf \anylong} & \undirected  & 
\multicolumn{1}{c|}{\anylong} & \undirected & \\ 
\hline \hline Obama & 307,985 & 60,137,108 & 19,204,843 & 8,205,166 & 861,394 & 195.3 & 124.7 & 26.6 & 5.6 & 4,873,711\\
\hline Sarah Palin & 121,910 & 14,318,290 & 4,278,903 & 3,764,747 & 449,568 & 117.5 & 70.2 & 30.9 & 7.4 & 972,537\\
\hline Glenn Beck & 95,847 & 9,684,761 & 3,038,396 &  2,862,626 & 357,910 & 101.0 & 63.4 & 29.9 & 7.5 & 687,913\\
\hline Lakers & 76,926 & 4,668,618 & 949,194 & 1,030,722 & 91,436 & 60.7 & 24.7 & 13.4 & 2.4 & 301,558\\
\hline Fox News &114,530 & 17,197,997 & 5,497,221 & 3,889,892 & 462,306 & 150.2 & 96.0 & 34.0 & 8.1 & 1,231,519 \\
\hline
\end{tabular}
\normalsize
\end{table*}

Some initially plausible hypotheses are not consistent with our data.
For instance, 
one might think that sparsity or having a smaller relative amount of
labeled training data would affect the performance rankings.  
However, neither graph sparsity nor the relative or absolute
amount of 
users in the graph explain why there are more improvements in
``Glenn Beck'' than ``Lakers'' 
or why ``Fox News'' performs relatively poorly.
Table \ref{tb:sparsitystatistics} shows the average degree in different topics as an approximation for sparsity. 
In comparison to the Glenn
Beck graphs, the Lakers graphs 
are denser.
And, Fox News has the highest proportion of labeled to unlabeled data
(since it has the fewest users), but our algorithm yields relatively
few improvements there.

However, the topic statistics depicted back in
Figure~\ref{fig:sentimentStatistics} do reveal two important facts
that help explain why ``Lakers'' and ``Fox News'' act
differently. First, they are the two topics for which the mutual
\follow edges have the lowest probability of connecting same-label
users, which  explains the paucity of red $\ast$'s in those topics'
plots.  Second, the reason ``Lakers'' and ``Fox News'' exhibit more
statistically significant improvements for the @ graph is that, as
Figure  \ref{fig:sentimentStatistics} again shows, they are the topics
for which directed @ edges and \undirected @ edges, respectively,
have the highest probability among all edge types of corresponding to
a shared label.  Thus, we see that when the quality of the underlying
graph is high, our graph-based approach can produce significant
improvements even when the graph is quite sparse --- for Fox News,
there are only 5 
\undirected @ 
pairs. 
(The high performance of SVM Vote for ``Lakers'' makes it more difficult to make further improvements.)

\vpara{Variation in SVM training}  We now briefly mention our
experiments with two alternative training sets for the tweet-level SVM
that underlies the SVM vote baseline: (a) a single set of
out-of-domain tweets labeled using emoticons as distant supervision
\cite{Read:05a}; (b) the same 5 topical sets described in \S \ref{sec:expproc}, except that we discarded tweets to enforce a 50/50 class balance.  For (a), the statistical-significance results were roughly the same as for our main training scheme, except for ``Obama'', where the SVM-vote results themselves were very  poor. Presumably, a graph-based approach cannot help if it is based on extremely inaccurate information.  For (b), there were some small differences in which graphs provided significant improvements;  but we believe that in a semi-supervised setting, it is best to not discard parts of what little labeled data there is.

\vpara{Adding more \unlabeled data}
How much does adding more \unlabeled data help? To provide some insight
into this question, we consider one underlying 
graph type
and evaluation metric ---
 directed \follow graph, accuracy --- and plot in Figure
\ref{fig:unlabeledperf} how performance is affected by increasing the
number of \unlabeled users. 
Note that what we plot is the average accuracy for the largest connected
component of the \labeled evaluation data, since this constitutes a more stable measure
with respect to increase in overall graph size.  
Also, note that the way we increase the number of unlabeled users is 
taking them from the crawl 
we obtained in our initial pass over users, which contained
$1,414,340$ users, $1,414,211$ user profiles, $480,435,500$ tweets,
$274,644,047$  \follow edges, and  $58,387,964$ @-edges; 
Table \ref{tb:fullgraphstat} shows the statistics for all the 
expanded
 graphs we collected. 

Figure \ref{fig:unlabeledperf} shows that HGM-\srank{} is
generally better than the SVM Vote baseline and at worst does
comparably.  HGM-\sta\ tends to degrade more than HGM-\srank{},
suggesting that learning-based parameter estimation is effective at
adjusting for 
graphs
with more unlabeled data. 
Edge density does not explain the relatively larger improvements in ``Lakers'' and
``Fox News'' 
because those are not the densest graphs.

\section{Related work}
\label{sec:related}

Recently, there has been some work on sentiment
analysis on Twitter, focusing on the tweet level
\cite{Li+Hoi+Chang+Jain:10,Oconnor+Balasubramanyan+Routledge+Smith:10,barbosa:2010,davidov:2010,Jiang:acl11}.
 Of 
deployed twitter-sentiment websites
(
e.g., 
www.tweetfeel.com, www.tweetsentiments.com, www.twitrratr.com),
the techniques employed are generally standard 
tweet-level 
algorithms
that ignore links between users.

There has been some previous work on
automatically determining
 user-level opinion or ideology
\cite{Agrawal+al:03a,Thomas+Pang+Lee:06a,yu2008classifying,Mullen+Malouf:08a,Gryc:10},
generally looking at information just in the text that the users generate.

A number of different graphs have been exploited in document- or sentence-level sentiment analysis
\cite{Agrawal+al:03a,Pang+Lee:04a,Agarwal+Bhattacharyya:05a,Pang+Lee:05a,Mullen+Malouf:08a,Somasundaran+Namata+Getoor+Wiebe:09,Wu+Tan+Cheng:2009a,Tanev+pouliquen+Zavarella+Steinberger:10,Jiang:acl11}, including in a semi-supervised setting \cite{Goldberg+Zhu:06a,Goldberg+Zhu+Wright:07a,Sindhwani+Melville:2008a}. Our use of @-mentions is similar to previous sentiment-analysis work using the network of references that one speaker makes to another \cite{Thomas+Pang+Lee:06a}.

\section{Conclusions and future work}
\label{sec:conclude}

We demonstrated that user-level sentiment analysis can be
significantly improved by incorporating link information from a social
network.  These links can correspond to attention, such as
when a Twitter user wants
to pay attention to another's status updates, or homophily, where
people who know each other are connected.  Choice of  follower/followee
network vs @ network and directed vs. \undirected connections
represent different aspects of the homophily vs attention alternatives.
We have some slight evidence that considering both homophily and attention is
superior to homophily alone, although we also observed some exceptions.
Regardless, significant gains can be
achieved even when the underlying graph is very sparse, as long as
there is a strong correlation between user connectedness and shared
sentiment.

The general idea in this 
paper, to explore social network structures to help sentiment analysis, represents an interesting research direction in social network mining. 
There are many potential future directions
for this work.
A straightforward
task
would be to build a larger \labeled dataset across more general
topics;
also, datasets
from
other online social media systems with other kinds of social networks
and more information
on
users 
would also be worth exploring.
Looking farther ahead, different
models 
and semi-supervised learning algorithms 
for exploiting network structures
should
be beneficial. 
For example, we tried some preliminary experiments with a Markov
Random Field formulation, although the sparsity of the graphs may be
an issue
in applying such an approach.
Finding which parts of the whole network are helpful with respect to
prediction on a topic is 
another
interesting direction. 
Finally, building a theory of 
why and how users correlate on different topics in different kinds of
networks
 is an intriguing direction for future research.

\vpara{Acknowledgments} Portions of this work were
done while the first author was interning at Microsoft Research
Asia. We thank Claire Cardie, Cristian Danescu-Niculescu-Mizil, Jon Kleinberg, Myle Ott, Karthik Raman, Lu Wang, Bishan Yang, Ainur Yessenalina  
and the anonymous reviewers for helpful comments. This work is
supported by Chinese National Key Foundation Research
60933013 and 61035004, a Google Research Grant, a grant
from Microsoft, ONR
YIP-N000140910911, 
China's National High-tech R\&D Program 
2009AA01Z138, Natural Science Foundation of China
61073073, US NSF DMS-0808864 and IIS-0910664, and a Yahoo! Faculty Research and Engagement Award.

\small
\bibliographystyle{abbrvnat}

\end{document}